\documentclass[letterpaper, 10 pt, conference]{ieeeconf}  

\IEEEoverridecommandlockouts                              

\usepackage{cite}
\usepackage{amsmath,amssymb,amsfonts}
\usepackage{algorithmic}
\usepackage{graphicx}
\usepackage{textcomp}
\usepackage{xcolor}
\usepackage{tikz}
\usepackage{rotating}
\usepackage{booktabs}
\usepackage{multirow}
\usepackage{array}
\usepackage{url} 
\usepackage[hidelinks]{hyperref}
\usepackage{placeins}
\usetikzlibrary{positioning, shapes.geometric, arrows}

\title{\LARGE \bf
Unified Attention Modeling for Efficient Free-Viewing and Visual Search via Shared Representations
}

\author{Fatma Youssef Mohammed and Kostas Alexis
\thanks{This work was supported by Horizon Europe under Grant EC 101120732.}
\thanks{The authors are with the Department of Engineering Cybernetics, Norwegian University of Science and Technology (NTNU), Trondheim, Norway.
        {\tt\small fatma.y.m.a.e.f.mohammed@ntnu.no}}%
}

\begin{document}

\maketitle
\thispagestyle{empty}
\pagestyle{empty}

\begin{abstract}


Computational human attention modeling in free-viewing and task-specific settings is often studied separately, with limited exploration of whether a common representation exists between them. This work investigates this question and proposes a neural network architecture that builds upon the Human Attention transformer (HAT) to test the hypothesis. Our results demonstrate that free-viewing and visual search can efficiently share a common representation, allowing a model trained in free-viewing attention to transfer its knowledge to task-driven visual search with a performance drop of only 3.86\% in the predicted fixation scanpaths, measured by the semantic sequence score (SemSS) metric which reflects the similarity between predicted and human scanpaths. This transfer reduces computational costs by 92.29\% in terms of GFLOPs and 31.23\% in terms of trainable parameters. 


\end{abstract}

\section{Introduction}
\label{sec: Introduction}
Human visual attention is an information selection process which enables us to focus on specific ``salient'' regions of a scene while ignoring others, therefore minimizing cognitive load~\cite{itti2002model,gottlieb2018towards,carrasco2011visual}. Accordingly, the study and computational modeling of human visual attention~\cite{ungerleider2000mechanisms,tsotsos2021computational,borji2012state} are of great importance both in the domain of neuroscience as well as in terms of the host of relevant applications. For instance, in guided inspection, robotic systems can prioritize processing regions of human interest rather than uniformly allocating resources across all regions, thus enhancing efficiency in terms of time and resource utilization. 


To dive into the process of computational modeling of human visual attention, it is essential to consider the fact that humans perceive their field of view with varying resolution. The fovea, a small central area in the visual field, provides very high-resolution perception, which in turn diminishes progressively toward the periphery. Thus, to effectively gather information, humans have to make eye movements directing their fovea to the most interesting areas in the visual field. These eye movements consist both of a) fixations, where the gaze remains relatively stable on a particular point of interest, as well as b) saccades which are rapid eye movements shifting the gaze from one location to another~\cite{hoffman1995role}. 


A key question in modeling the process of human visual attention is the modulation of attention between focusing or not on a particular visual task. Broadly, human visual attention can be categorized into being ``Bottom-Up'' (task-free) and ``Top-Down'' (task-specific)~\cite{connor2004visual,katsuki2014bottom}. Bottom-up or ``free-viewing'' attention is driven by the salience of features in the visual input, where certain elements stand out due to their inherent properties relative to the background and how those correlate with intrinsic properties of the human attention process. Free-viewing is considered to have its roots in evolutionary processes. In contrast, top-down or ``task-specific'' attention is guided by particular goals and tasks and thus it focuses on diverse entities in the visual stimuli. 


Most studies focus on predicting the human attention in either free-viewing or task-specific settings (with ``visual search'' being the most common task). A recent model, the Human Attention Transformer (HAT)~\cite{yang2024unifying}, was proposed as a unified approach capable of predicting fixation scanpaths for both the visual search and free-viewing tasks without requiring architectural changes. As demonstrated, the same architecture can be trained for one task and retrained for another. Despite HAT's ability to architecturally handle both tasks, for all of its components beyond an early attention-agnostic ResNet-50 based encoder it still requires retraining from scratch when transitioning between free-viewing and visual search and does not address whether a common representation exists between the two tasks. 

Departing from these limitations, this work aims to investigate the question of a deeply common representation between free-viewing and task-specific visual search by building on the HAT model, and enabling the identification of a shared representation between both tasks. By leveraging this shared representation, it is possible to use the features learned from free-viewing as input for the other task without the need for retraining, thus improving computational efficiency and reducing training costs. Extensive numerical evaluations allow to demonstrate the performance of our approach. Thus, this work paves the way for the deployment of models that can predict both visual attention modalities with the overwhelming amount of parameters and computations shared. This, in turn, can enable high-performance architectures that are simultaneously parsimonious, especially regarding how they scale to multiple visual information-gathering tasks. 

The remainder of this paper is structured as follows. Section~\ref{sec: Related Work} presents related work, while the proposed approach is detailed in Section~\ref{sec: Proposed Approach}. Results are presented in Section~\ref{sec: Results}, and conclusions are drawn in Section~\ref{sec: Conclusions}.

\section{Related Work}
\label{sec: Related Work}

One of the earliest bottom-up attention models was introduced by Itti and Koch~\cite{itti2002model}. Their model decomposes visual input into topographic feature maps based on color, intensity, and orientation. Center-surround differences are computed to create conspicuity maps, which are then combined into a saliency map. Their contribution greatly influenced the field and led to large family of methods on visual saliency modeling~\cite{tsotsos2021computational,borji2012state}. However, this purely stimulus-driven approach does not account for the influence of task goals on attention.

Later models incorporated task-driven components to enhance attention modeling. In \cite{navalpakkam2005modeling}, a framework integrates top-down relevance with bottom-up salience by identifying task-relevant entities using hand-crafted, non-learnable prior knowledge and biasing attention toward such target features. This approach aligns with neurobiological evidence of multiple attention-related maps in the brain~\cite{colby1999space,gottlieb1998representation,kustov1996shared}. Analogously, the Contextual Guidance Model~\cite{torralba2006contextual} also integrates both bottom-up and top-down influences thus positing that fixation selection is driven by a combination of bottom-up salience and contextual priors. 



A more detailed breakdown of attention guidance was provided in~\cite{wolfe2017five}, where a framework was proposed for five factors guiding attention in visual search: bottom-up saliency, top-down saliency, scene-based guidance, value-based guidance and history-based guidance. The framework indicates that purely bottom-up models often struggle to accurately predict eye fixations, even in free-viewing scenarios, unless they incorporate the structural arrangement of objects in a scene.

With the advent of deep learning, modern approaches leverage deep features to predict saliency maps and fixation scanpaths (the sequence of fixations a person performs while viewing an image)~\cite{kummerer2022deepgaze,sun2019visual,zelinsky2021predicting,mondal2023gazeformer}. However, these models have largely been developed for either task-specific (top-down) or task-free (bottom-up) attention, without a unified approach to both. Recently, the Human Attention Transformer (HAT)~\cite{yang2024unifying} was introduced as a model capable of predicting fixation scanpaths for both visual search and free-viewing without requiring architectural modifications. The model uses a ResNet-50 based encoder trained agnostically to visual attention modeling (trained on panoptic segmentation) as a feature extractor and then
proposes a neural network architecture that has shown to be able to be adopted for both free-viewing and task-specific visual search, albeit with separate training of all other parameters. It represents the current state-of-the-art in human attention modeling. Unlike other deep learning models that struggle to generalize when retrained on a different attention type, HAT exhibits the capacity to adapt to both top-down and bottom-up attention, suggesting it captures essential features for both tasks. However, despite its flexibility, no explicit study has been conducted to analyze whether a common representation exists between the two tasks within HAT.

Motivated by prior work in modeling human visual attention and the recent advancements in deep learning, this work aims to investigate whether there is a shared representation between task-specific and task-free attention. Identifying a shared representation between the two would offer deeper insight into human attention mechanisms while improving computational efficiency. A unified model could generalize across both tasks, reducing the need for separate training and significantly reducing computational costs.

\section{Proposed Approach}
\label{sec: Proposed Approach}

\subsection{Datasets}
\label{sec: Datasets}
This study utilizes two publicly available datasets: \textbf{COCO-Search18}~\cite{chen2021coco} and \textbf{COCO-FreeView}~\cite{chen2022characterizing}. Both datasets have the same 6202 images, but differ in their task annotations. 

\begin{itemize}
    \item \textbf{COCO-FreeView} contains human fixation annotations collected during free-viewing, where participants explored the images without any specific task.
   \item \textbf{COCO-Search18} contains fixation sequences collected during a goal-driven visual search task. For each image, a specific object category was assigned as a target, and participants were asked to locate it. The dataset includes both \textit{target-present} and \textit{target-absent} conditions, with 3101 images in each subset.

\end{itemize}

In this work, we focus exclusively on the \textit{target-present} condition, where the target object is present in the image and human fixations reflect active search behavior. For brevity, we refer to this condition as \textbf{visual search} throughout the paper.

These datasets provide an opportunity to explore the representational overlap between task-free (free-viewing) and task-specific (visual search) attention, as they contain the same images annotated under different attention paradigms. To the best of our knowledge, this is the only dataset pair that provides both free-viewing and visual search fixations on the same images with distinct task annotations, making it valuable for studying shared representations across attention tasks.

\subsection{Network Architecture}
\label{sec: Network Architecture}
The proposed network architecture is inspired by the Human Attention Transformer (HAT) model~\cite{yang2024unifying}.
HAT consists of four main modules: \textit{feature extraction}, \textit{foveation}, \textit{aggregation}, and \textit{fixation prediction}.  

First, the input RGB image of size $(H, W)$ is processed by the \textbf{feature extraction module}, which includes a \textit{pixel encoder} (ResNet-50)~\cite{he2016deep} that extracts high-level semantic features and a \textit{pixel decoder} (MSDeformAtten)~\cite{zhu2020deformable} composed of $6$ transformer layers and $8$ attention heads that output multi-scale feature maps with four different resolutions. The lowest resolution is $(H/32, W/32)$, while the highest is $(H/4, W/4)$. 
Both pixel encoder and decoder are initialized with the COCO pretrained weights for panoptic segmentation from~\cite{cheng2022masked} while the pixel encoder weights are kept fixed during training. Reusing the pixel encoder (ResNet-50) weights learned from panoptic segmentation enables the model to have a universal feature extractor that is capable of extracting object-level features. These features can then be tailored for the fixation prediction task for either free-viewing or visual search by training the pixel decoder for either task.

\begin{figure*}[h!] 
    \centering
    \includegraphics[width=\linewidth]{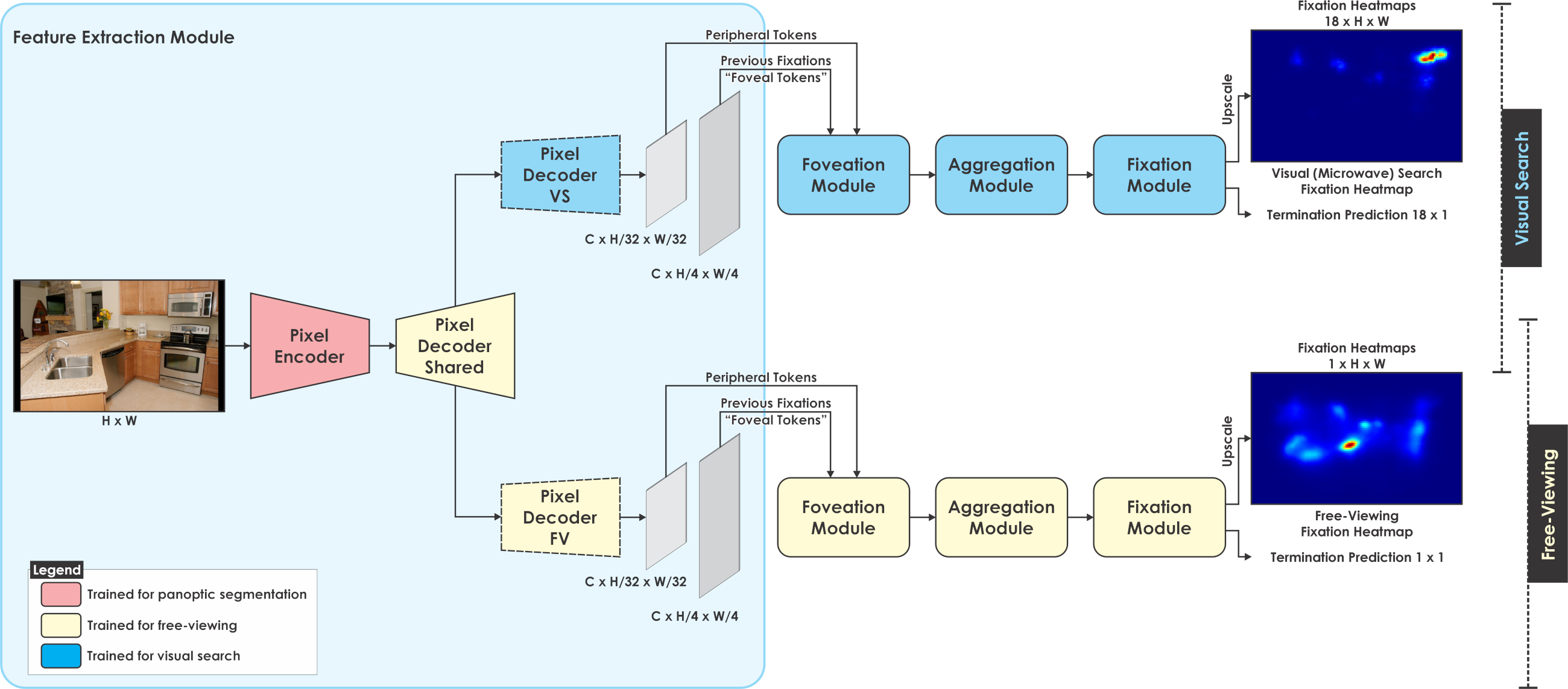} 
    \caption{Proposed architecture for predicting human attention in both the free-viewing and visual search tasks building upon and extending HAT~\cite{yang2024unifying}. The pixel encoder is ResNet-50~\cite{he2016deep}, while the pixel decoder is MSDeformAttn~\cite{zhu2020deformable}. The total number of transformer layers in the pixel decoder is 6, where the number of shared layers can range from 1 as in Early-Split 1-5 (ES$_{1,5}$) to 6 layers as in Late-Split (LS), leading to 5 task-specific layers in ES$_{1,5}$ and no task-specific layers in LS.}
    \label{fig:FV and TP Architecture}
\end{figure*}

The lowest and highest resolution feature maps output of the pixel decoder are then passed to the \textbf{foveation module}, where fixation-related tokens are extracted. Previous fixation tokens are taken from the high-resolution feature map, resembling \textit{foveal vision}, while peripheral tokens are extracted from the low-resolution feature map, resembling \textit{peripheral vision}. These tokens are then fed into a transformer encoder of $3$ layers and $4$ attention heads, which serves as memory and updates its state with each new fixation.  

The output of the transformer encoder is passed to the \textbf{aggregation module}, which contains a transformer decoder composed of $6$ layers and $4$ attention heads that integrates information from memory using \textit{learnable task queries} \( Q \). For the \textit{visual search} task, \( Q = 18 \) (corresponding to 18 different objects), while for the \textit{free-viewing} task, \( Q = 1 \). 

The updated queries are finally passed to the \textbf{fixation prediction module}, which combines them with the high-resolution feature map from the feature extraction module by first passing the learned queries to a Multi-Layer Perceptron (MLP) of two layers and then convolving the resulted embedding with the high-resolution feature map to predict the current fixation heatmap. To predict the termination probability, which indicates whether the current fixation is the final one in the sequence, each learned query (corresponding to each task) is fed to a linear layer followed by a sigmoid activation function.

\subsection{Shared Representations for Unified Attention Modeling}

Unlike HAT~\cite{yang2024unifying}, which investigates separately the problem of free-viewing and visual search performance, in this work, we investigate the potential of shared computational modeling. To determine whether a common representation exists between visual search and free-viewing, the outlined model is adapted to predict human attention for both tasks simultaneously. Since each task has a different sequence of previous fixations, the split between the two tasks must occur within the \textbf{feature extraction module}, either at the output of the pixel decoder  or within the pixel decoder (making the model more customized for visual search), since the weights of the pixel encoder are fixed.

Figure~\ref{fig:FV and TP Architecture} illustrates the newly proposed architecture, with the splitting stage occurring in the feature extraction module, where a number of pixel decoder layers can be shared while the other can be task-specific, allowing for adaptation to a certain task while having a shared representation. The two distinct subsets of pixel decoder layers are labeled as ``Pixel Decoder VS'' and ``Pixel Decoder FV'' for the visual search and free-viewing branches, respectively. These layers are enclosed in dashed boxes to indicate that, in scenarios where all pixel decoder layers are shared, no task-specific layers are present.

The training scheme for the proposed architecture consists of two main stages. First, the free-viewing branch of the network is trained while keeping the weights of the visual search branch fixed. In the second stage, the shared layers are frozen, utilizing the trained weights from the free-viewing branch, and the visual search branch is then trained. This strategy assumes that there is interplay between bottom-up and top-down visual attention~\cite{katsuki2014bottom,connor2004visual} a question still under investigation in the neuroscience community. It further hypothesizes that as bottom-up attention is task- and object-agnostic, it can offer quite generic features within early stages of its representation, which can then be repurposed for task-driven attention. The performance of this approach is compared with an end-to-end training scheme on the visual search branch (i.e., assuming fully independent training of the top-down visual search task such that it exploits all the expressiveness of the neural network). It is noted that by end-to-end here we refer to the training of all parameters in HAT for the visual search task except the pixel encoder, which utilizes weights from the pretrained ResNet-50. For simplicity we shall call this ``end-to-end'' training of HAT throughout this document.

If a common representation exists between the two tasks, then the features learned from the free-viewing task should serve as input for visual search attention prediction without significant performance degradation. This would allow for a significant reduction in training computational costs by reusing shared layers instead of retraining them from scratch. Our goal is to identify these shared representations, as a more generalizable representation results in greater computational efficiency.

\subsection{Investigation Details}\label{sec:investigation}

To investigate if there is an interaction between bottom-up and top-down attention and explore the value of shared representations, six different possible shared representation layers in the pixel decoder (depending on how many layers of the pixel decoder layers are kept fixed) are explored and compared with the HAT model. Specifically, we have: 
\begin{enumerate}
    \item \textbf{Late-Split} (LS): All six transformer layers are shared.
    \item \textbf{Early-Split 5-1} (ES$_{5,1}$): Five shared layers and one task-specific layers.
    \item \textbf{Early-Split 4-2} (ES$_{4,2}$): Four shared layers and two task-specific layers.
    \item \textbf{Early-Split 3-3} (ES$_{3,3}$): Three shared layers and three task-specific layers.
    \item \textbf{Early-Split 2-4} (ES$_{2,4}$): Two shared layers and four task-specific layers.
    \item \textbf{Early-Split 1-5} (ES$_{1,5}$): One shared layers and five task-specific layers.
\end{enumerate}
The main difference between Late-Split (LS) and all Early-Split (ES) variants is that with LS, the whole feature extraction module is reused for visual search attention prediction. In the case of ES variants, part of the pixel decoder is trained for visual search, allowing the network to be more customized for visual search which can improve its performance on that task while increasing the computational costs for training.

\subsection{Training}
\label{sec: Training}
The same training hyperparamters were utilized as in HAT~\cite{yang2024unifying}. A combination of focal loss and binary cross entropy loss was used for fixation heatmap prediction and termination probability prediction, receptively. The AdamW optimizer was used with a fixed learning rate of 0.0001.  A batch size of 32 was used in all experiments. The training was conducted for only 15 epochs in contrast to the 30 epochs used in HAT since part of the pixel decoder is initialized with HAT free-viewing pretrained weights and kept fixed during training, reducing the total number of the parameters to be learned and leading to faster convergence.


\section{Results}
\label {sec: Results}

The utilized metrics, alongside a detailed analysis of our results, are presented in this Section.

\subsection{Evaluation Metrics}

Two sets of metrics were used to evaluate the model’s prediction performance, following HAT~\cite{yang2024unifying}. The first set assesses the similarity between the predicted and human scanpaths using \textbf{Sequence Score (SS)} and \textbf{Semantic Sequence Score (SemSS)}. In SS, scanpaths are transformed into sequences of fixation clusters and compared using a string-matching algorithm~\cite{borji2013analysis}. SemSS extends this by converting scanpaths into sequences of semantic labels (object category using COCO segmentation annotation) representing the fixated pixels~\cite{yang2022target}.

The second set of metrics focuses on conditional saliency, evaluating how well the model’s current prediction is given previous ground truth fixations. These include \textbf{Information Gain (cIG)}, \textbf{Normalized Scanpath Saliency (cNSS)}, and \textbf{Area Under the Curve (cAUC)}. cIG measures the difference in the average log-likelihood between the model and a baseline model~\cite{kummerer2015information,kummerer2021state}. The baseline is represented by a fixation density map for each object visual search. The fixation density map is calculated by averaging the fixation maps (smoothed by a gaussian kernel) for all training fixations of the same object~\cite{yang2024unifying}. cNSS quantifies the correspondence between the predicted saliency map and human scanpaths by computing the mean of normalized salience values at fixation locations along a subject’s scanpath~\cite{peters2005components,kummerer2021state}. cAUC treats the model’s saliency map as a binary classifier on each pixel, labeling pixels with saliency values above a threshold as fixated and the rest as non-fixated. Human fixations serve as ground truth. By varying the threshold, a Receiver Operating Characteristic (ROC) curve is generated, with the area under this curve indicating how well the saliency map predicts actual human fixations~\cite{wilming2011measures,kummerer2021state}.

\subsection{Analysis of Results}

First, it is noted that to ensure a fair comparison, instead of retraining the free-viewing HAT model, the pretrained weights provided by HAT~\cite{yang2024unifying} on the free-viewing task were used in all experiments.

To clearly examine the capability of building on top of free-viewing (bottom-up attention) and predicting fixations in visual search (top-down attention), the LS network shared configuration (Section~\ref{sec:investigation}) was used. In this case, the complete feature extraction module trained for free-viewing was utilized ``as is'' by the visual search branch (i.e., without further task-specific training).

To further examine the optimal sharing configuration to achieve the best performance on the visual search task with minimal additional training cost, different Early-Split configurations were also considered. Within the different ES configuration, the relevant shared layers of the pixel decoder (Section~\ref{sec:investigation}) were also fixed on the pretrained HAT model on free-viewing.

The LS configuration achieves comparable results with the fully trained HAT model on visual search with the cNSS metric exhibiting the most significant performance drop of 7.57\%. This highlights the ability to have a largely unified shared network that predicts attention in both free-viewing and visual search tasks simultaneously, while achieving the state-of-the-art results for free-viewing as reported by HAT~\cite{yang2024unifying} and a minor performance drop in visual search compared to the state-of-the-art. Training at least one of the pixel decoder layers (Early-Split configuration) on the visual task yields better performance than the LS configuration at the cost of increasing the number of training parameters and computational cost. The performance comparison between LS and different ES configurations against the end-to-end trained HAT model on visual search is summarized in Table~\ref{tab: comparison between HAT and our model using pretrained HAT weights}. The best (maximum) values for each metric are highlighted in bold.


In the case of LS configuration, when compared to the HAT model trained for visual search, the shared configuration reduces the computational cost in terms of GFLOPS by 92.29\% and the number of training parameters of visual search by 31.23\%. Notably, in HAT,  the authors only freeze the pixel encoder from the beginning of the training and utilize the learned weights on panoptic segmentation on the COCO dataset (i.e., agnostic to attention modeling). This reduces the number of trainable parameters from 42.783 to 19.328 million and the computational cost from 38.349 to 24.931 GFLOPS. Unlike HAT, here we demonstrate that pixel decoder layers trained for free-viewing can be repurposed for visual search and extend the HAT paradigm to additionally freeze the pixel decoder and reuse its pretrained weights on free-viewing. We thus achieve a further significant reduction in computational cost and training parameters with on-par performance with HAT. The number of training parameters and the computational cost (in terms of GFLOPs) per module is shown in Table~\ref{tab:model_stats}. 
Table~\ref{tab:FLOPS_comparison compared to HAT} provide a detailed comparison of the reduction in computational cost and trainable parameters in LS and all variants of ES relative to HAT (with the pixel encoder fixed).


The results indicate that a shared representation between free-viewing and visual search is plausible and efficient, enabling the features extracted from a model trained on free-viewing to serve as input to the visual search task without significantly affecting its performance while saving major computational costs.

\begin{table}[h]
\caption{Performance Comparison between the HAT model and variants of the proposed approach reusing selected layers from the pre-trained HAT model (trained on free-viewing).}
\centering
\begin{tabular}{lcccccc}
\toprule
\textbf{Method} & \textbf{SemSS} & \textbf{SS} & \textbf{cIG} & \textbf{cNSS} & \textbf{cAUC} \\
\midrule
\textbf{HAT} & \textbf{0.543} & \textbf{0.470} & 2.399 & 5.086 & 0.977 \\
\textbf{LS} & 0.522 & 0.462 & 2.248 & 4.701 & 0.975 \\
\textbf{ES$_{5,1}$} & 0.532 & 0.466 & 2.415 & 5.157 & \textbf{0.979} \\
\textbf{ES$_{4,2}$} & 0.525 & 0.461 & 2.439 & 5.098 &0.978 \\
\textbf{ES$_{3,3}$} & 0.516 & 0.462 & 2.386 & 4.959 &  0.975\\
\textbf{ES$_{2,4}$} & 0.525 & 0.465 & \textbf{2.514} & \textbf{5.285} & 0.978 \\
\textbf{ES$_{1,5}$} & 0.525 & 0.463 & 2.409 & 	5.034 & 0.977 \\

\bottomrule
\end{tabular}
\label{tab: comparison between HAT and our model using pretrained HAT weights}
\end{table}


\begin{table}[!h]
\centering
\caption{Number of Parameters and Computational Cost per Module.}
\begin{tabular}{lcc}
\toprule
\textbf{Module/Component} & \textbf{Number of Parameters (M)} & \textbf{GFLOPS} \\
\midrule
Pixel Encoder        & 23.455 & 13.418 \\
Pixel Decoder        & 6.036  & 22.997 \\
Foveation            & 3.063  & 1.545  \\
Aggregation           & 9.489  & 0.376  \\
Fixation Prediction   & 0.740  & 0.013 \\
\bottomrule
\end{tabular}
\label{tab:model_stats}
\end{table}

\begin{table}[h]
\caption{Percentage of reduction in model size and computational cost in shared layers configuration compared to HAT.}
\centering
\begin{tabular}{lcc}
\toprule
\textbf{Method} & \textbf{Reduced Trainable Parameters\%} & \textbf{Shared FLOPS\%} \\
\midrule
\textbf{LS} & 31.23\% & 92.29\%  \\
\textbf{ES$_{5,1}$} & 24.05\% &  52.48\% \\
\textbf{ES$_{4,2}$} & 20.26\% & 39.69\% \\
\textbf{ES$_{3,3}$} &  16.47\% & 26.91\% \\
\textbf{ES$_{2,4}$} & 12.68\% & 14.12\% \\
\textbf{ES$_{1,5}$} &  8.89\% &  1.34\% \\
\bottomrule
\end{tabular}
\label{tab:FLOPS_comparison compared to HAT}
\end{table}

\begin{table}[!h]
\centering
\caption{Performance Comparison on the newly collected dataset between the HAT model and variants of the proposed approach reusing selected layers from the pre-trained HAT model (trained on free-viewing).}
\begin{tabular}{lccc}
\toprule
\textbf{Method} & \textbf{cIG} & \textbf{cNSS} & \textbf{cAUC} \\
\midrule
\textbf{HAT} & -0.295 & 2.151 & 0.937 \\
\textbf{LS} & -0.301 & 2.041 & 0.939 \\
\textbf{ES$_{5,1}$} & -0.172 & \textbf{2.198} & 0.944 \\
\textbf{ES$_{4,2}$} & \textbf{-0.116} & 2.137 & \textbf{0.946} \\
\textbf{ES$_{3,3}$} & -0.697 & 1.827 & 0.929 \\
\textbf{ES$_{2,4}$} & -0.447 & 1.954 & 0.938 \\
\textbf{ES$_{1,5}$} & -0.386 & 2.090 & 0.938 \\
\bottomrule
\end{tabular}
\label{tab: comparison on the new dataset between HAT and our model using pretrained HAT weights}
\end{table}

\begin{figure*}[h]
    \centering
    \setlength{\tabcolsep}{4pt} 
    \begin{tabular}{ccccc}
        & \textbf{Car} & \textbf{Stop Sign} & \textbf{Bottle} & \textbf{Microwave} \\ 
        \centering \textbf{LS} & 
        \raisebox{-0.5\height}{\includegraphics[width=0.22\textwidth]{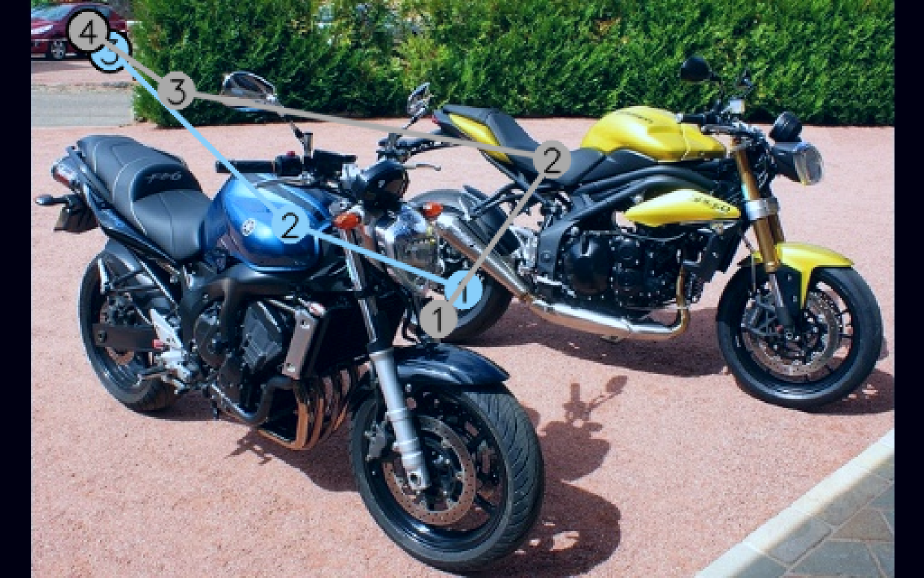}} & 
        \raisebox{-0.5\height}{\includegraphics[width=0.22\textwidth]{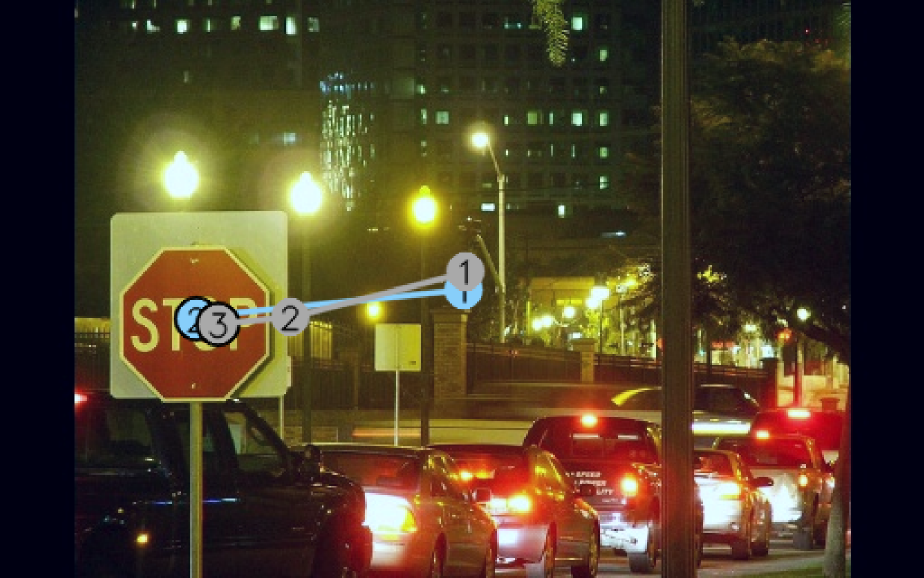}} & 
        \raisebox{-0.5\height}{\includegraphics[width=0.22\textwidth]{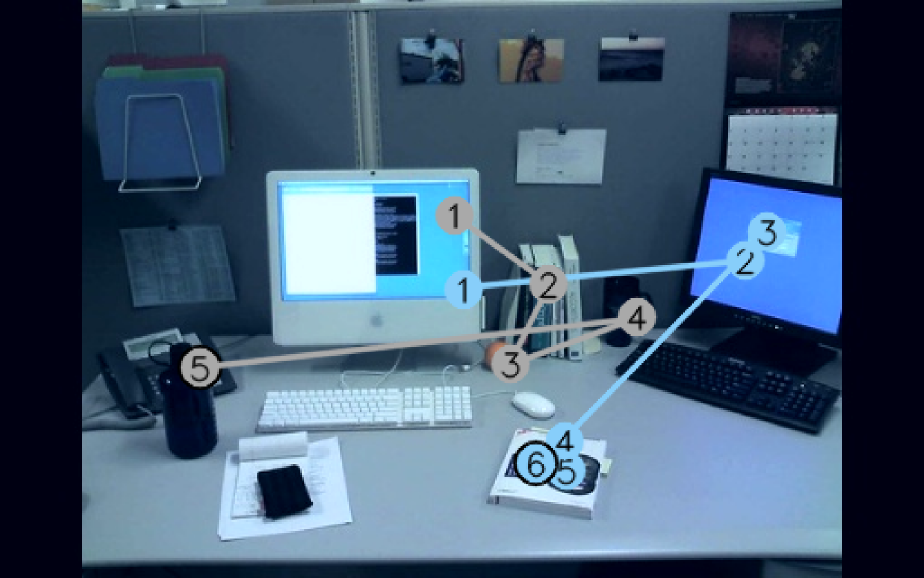}} & 
        \raisebox{-0.5\height}{\includegraphics[width=0.22\textwidth]{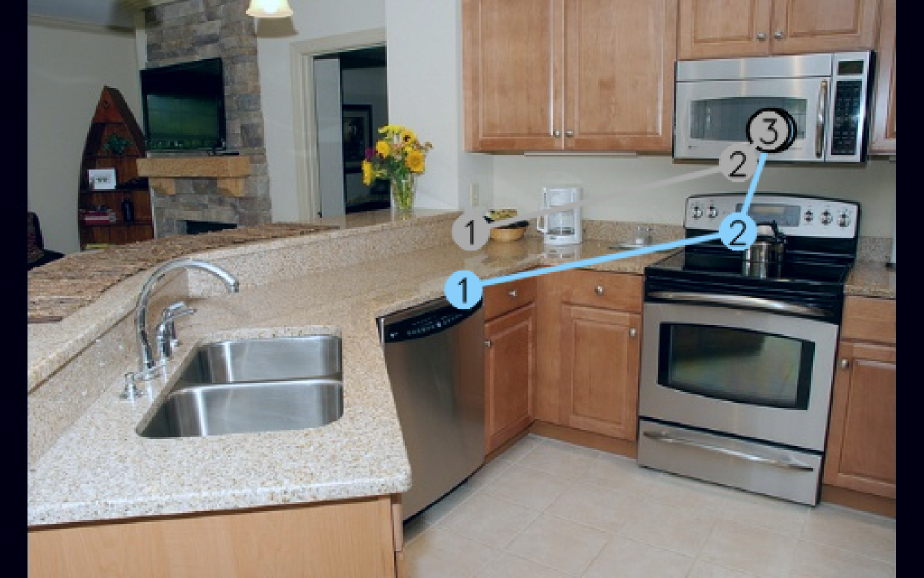}} \\ \\
        \centering \textbf{ES$_{5,1}$ }& 
        \raisebox{-0.5\height}{\includegraphics[width=0.22\textwidth]{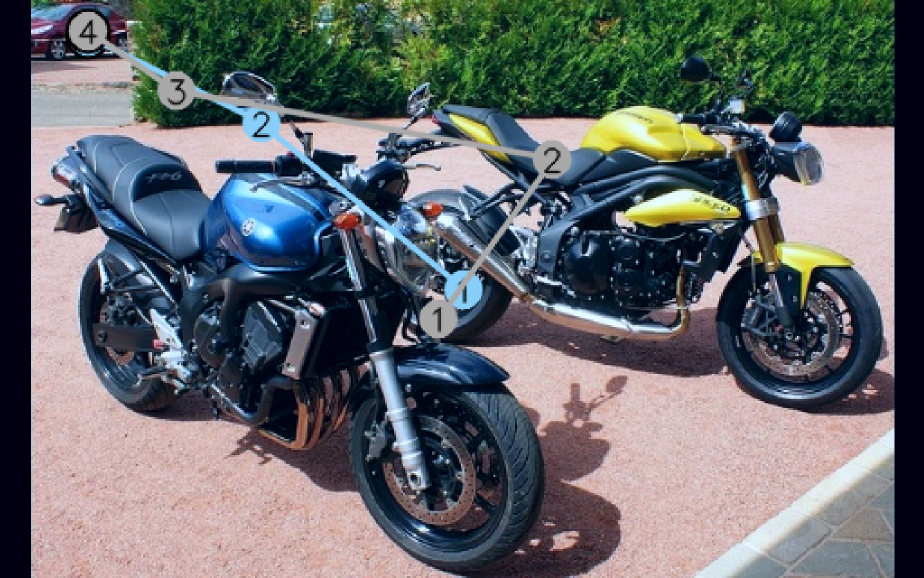}} & 
        \raisebox{-0.5\height}{\includegraphics[width=0.22\textwidth]{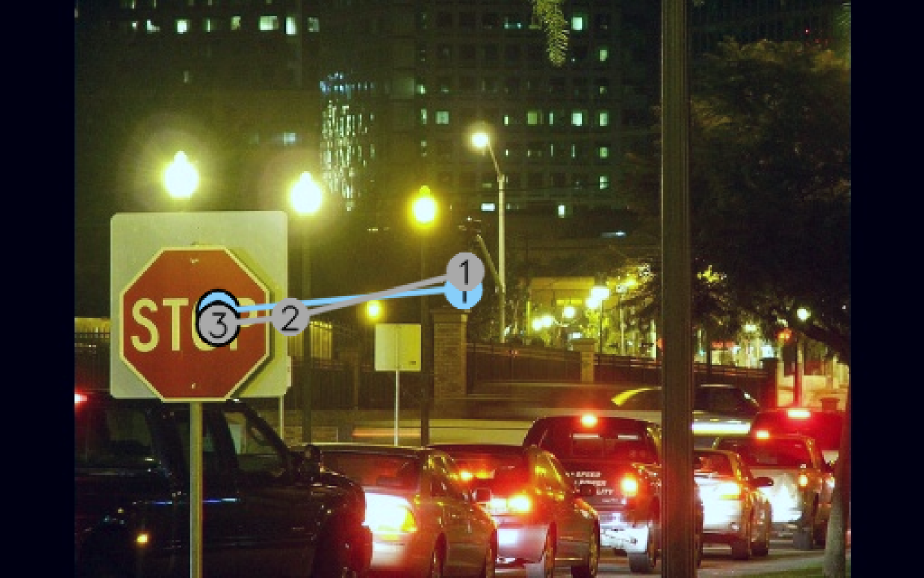}} & 
        \raisebox{-0.5\height}{\includegraphics[width=0.22\textwidth]{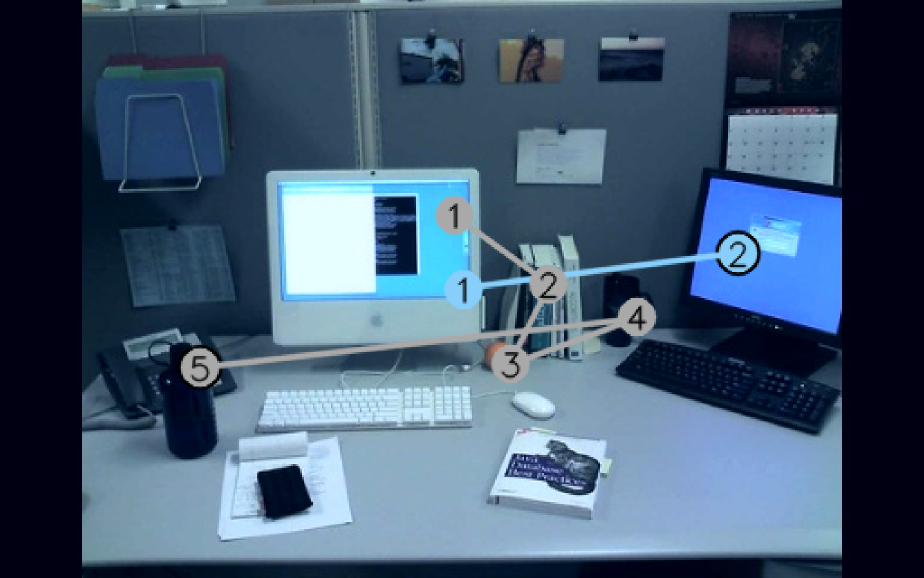}} & 
        \raisebox{-0.5\height}{\includegraphics[width=0.22\textwidth]{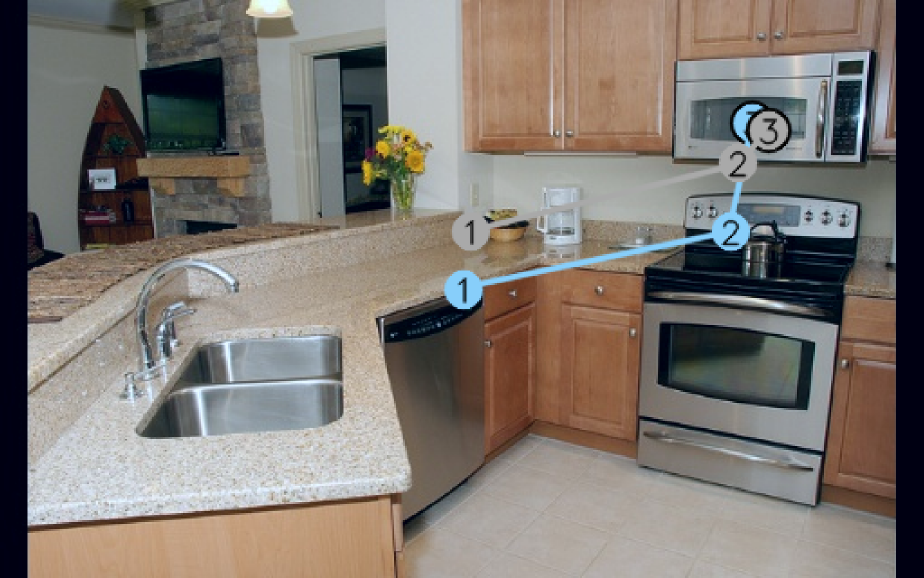}} \\ \\
        \centering \textbf{ES$_{4,2}$ } & 
        \raisebox{-0.5\height}{\includegraphics[width=0.22\textwidth]{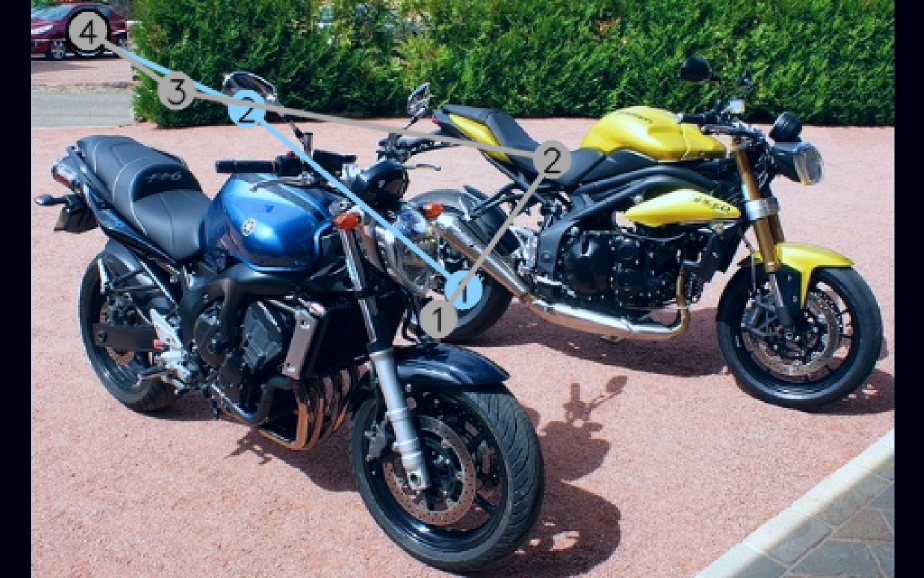}} & 
        \raisebox{-0.5\height}{\includegraphics[width=0.22\textwidth]{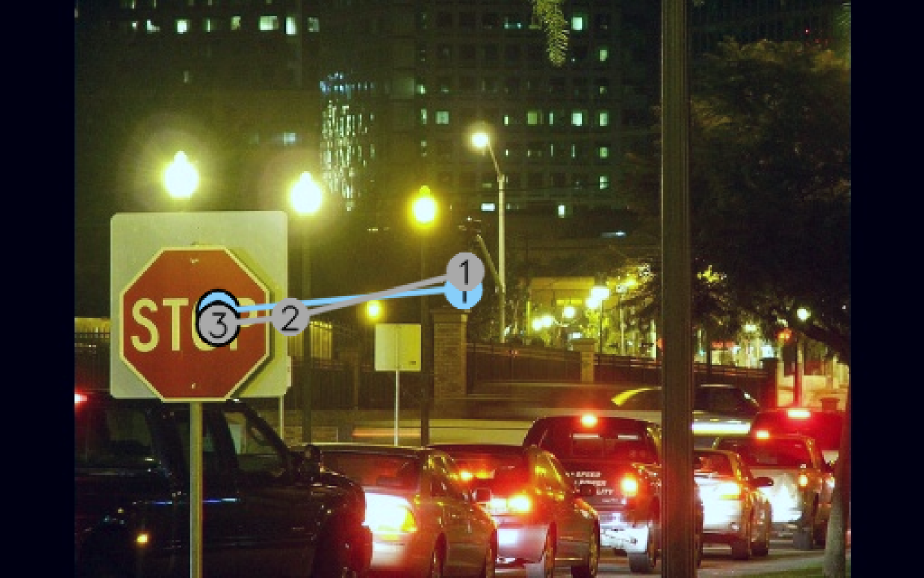}} & 
        \raisebox{-0.5\height}{\includegraphics[width=0.22\textwidth]{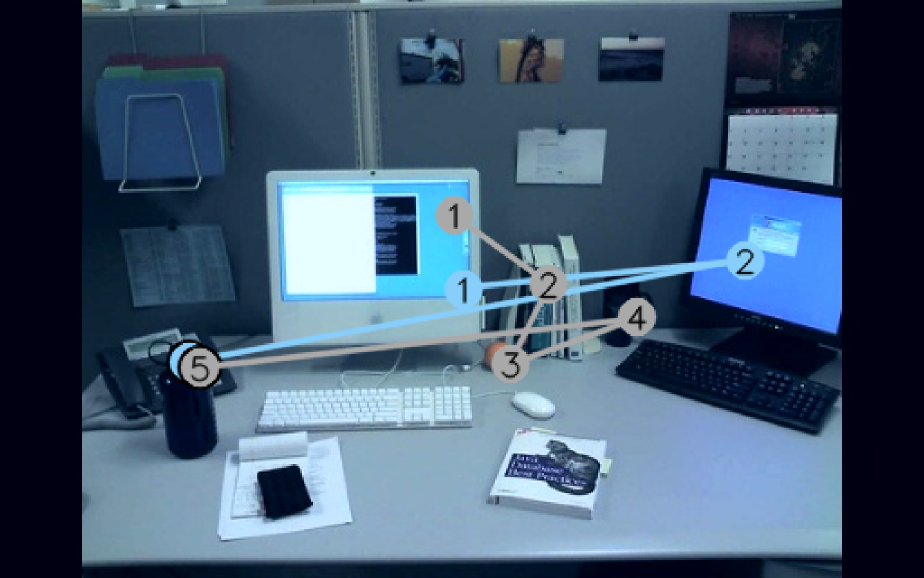}} & 
        \raisebox{-0.5\height}{\includegraphics[width=0.22\textwidth]{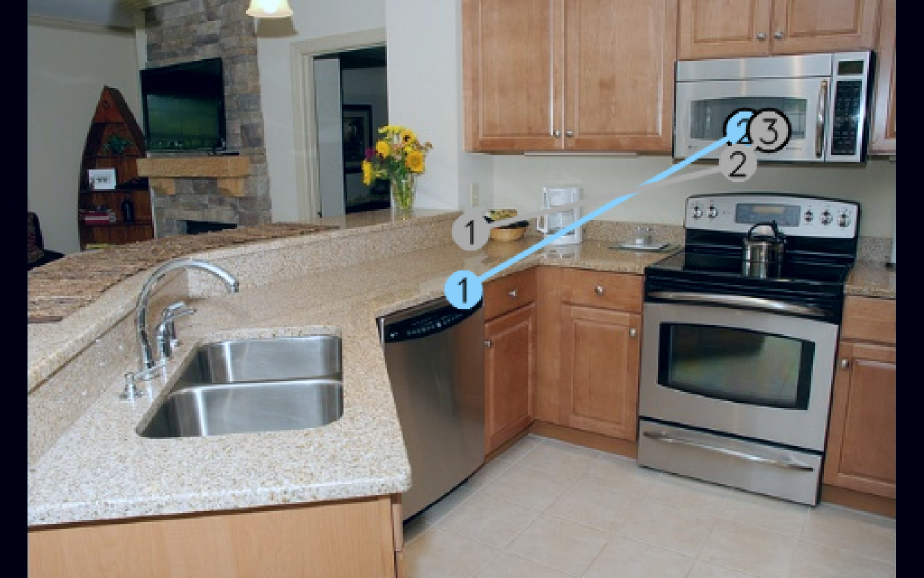}} \\ \\
        \centering \textbf{ES$_{3,3}$ } & 
        \raisebox{-0.5\height}{\includegraphics[width=0.22\textwidth]{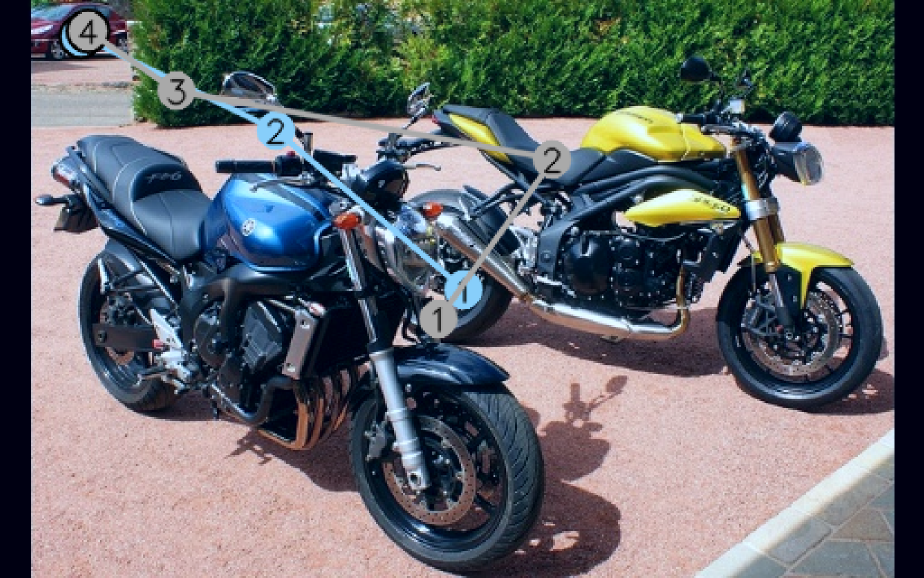}} & 
        \raisebox{-0.5\height}{\includegraphics[width=0.22\textwidth]{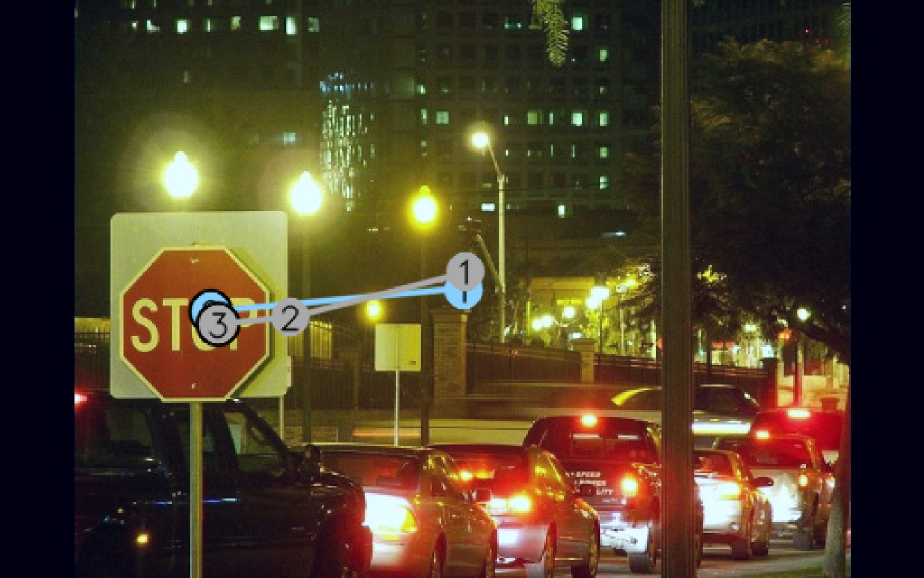}} & 
        \raisebox{-0.5\height}{\includegraphics[width=0.22\textwidth]{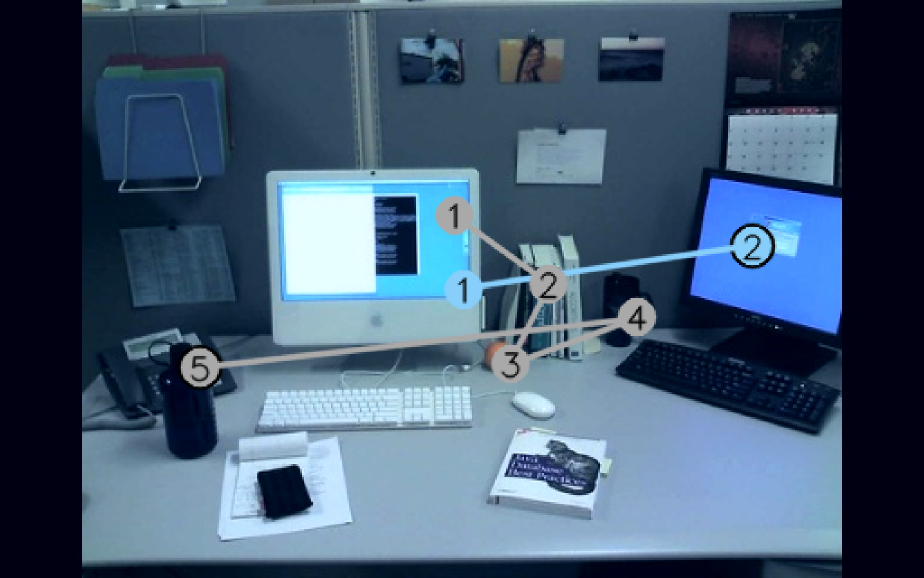}} & 
        \raisebox{-0.5\height}{\includegraphics[width=0.22\textwidth]{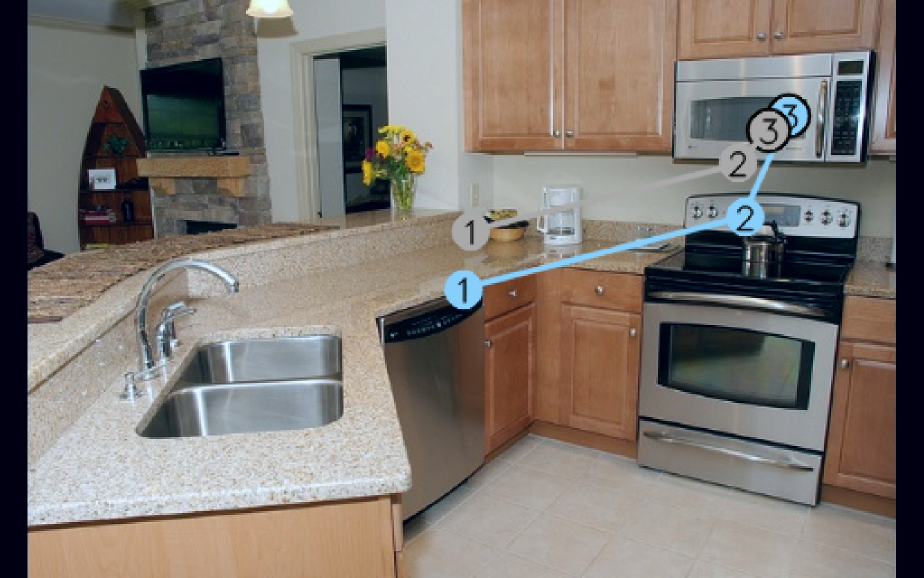}} \\ \\
        \centering \textbf{ES$_{2,4}$ } & 
        \raisebox{-0.5\height}{\includegraphics[width=0.22\textwidth]{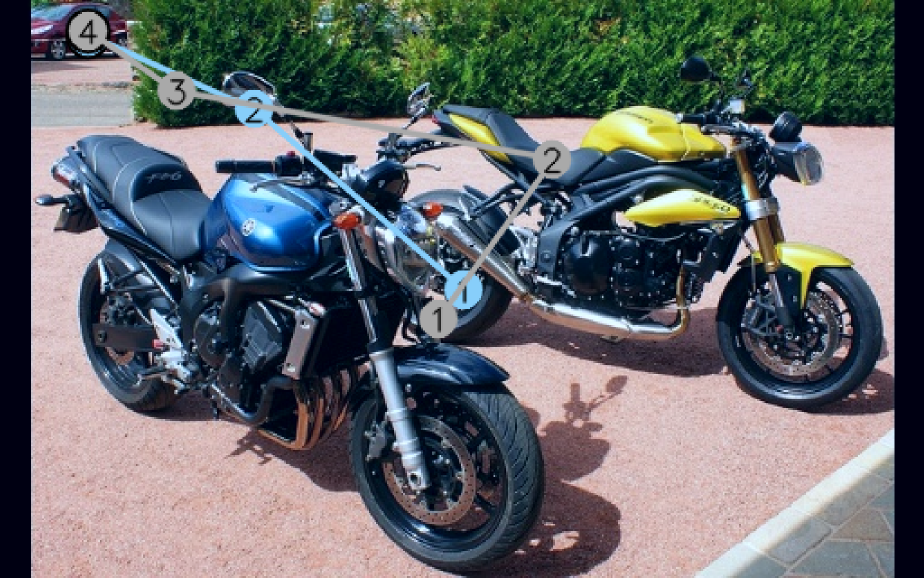}} & 
        \raisebox{-0.5\height}{\includegraphics[width=0.22\textwidth]{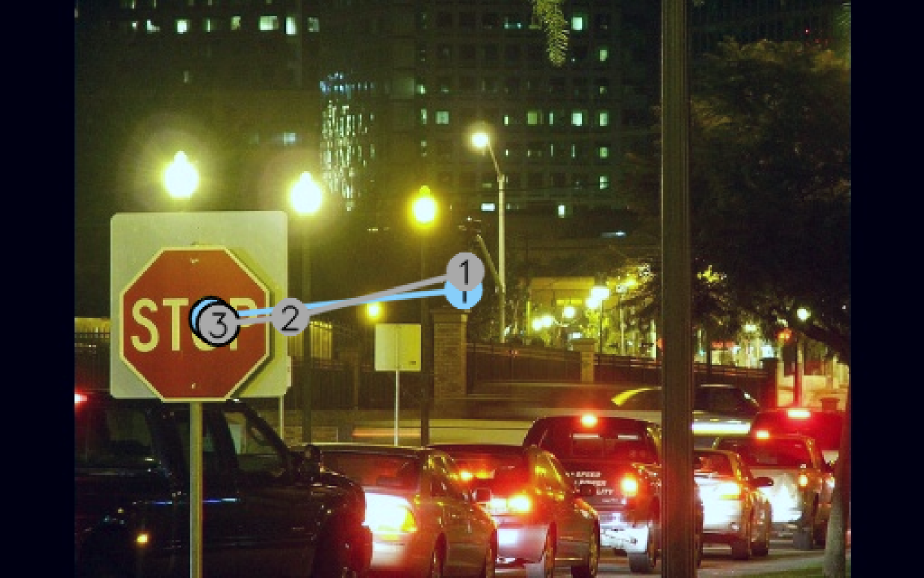} }& 
        \raisebox{-0.5\height}{\includegraphics[width=0.22\textwidth]{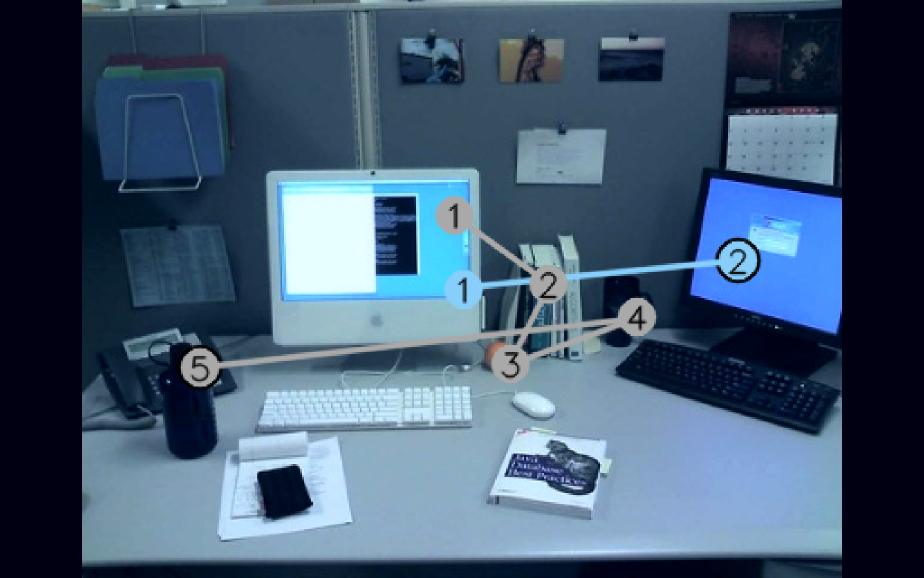}} & 
        \raisebox{-0.5\height}{\includegraphics[width=0.22\textwidth]{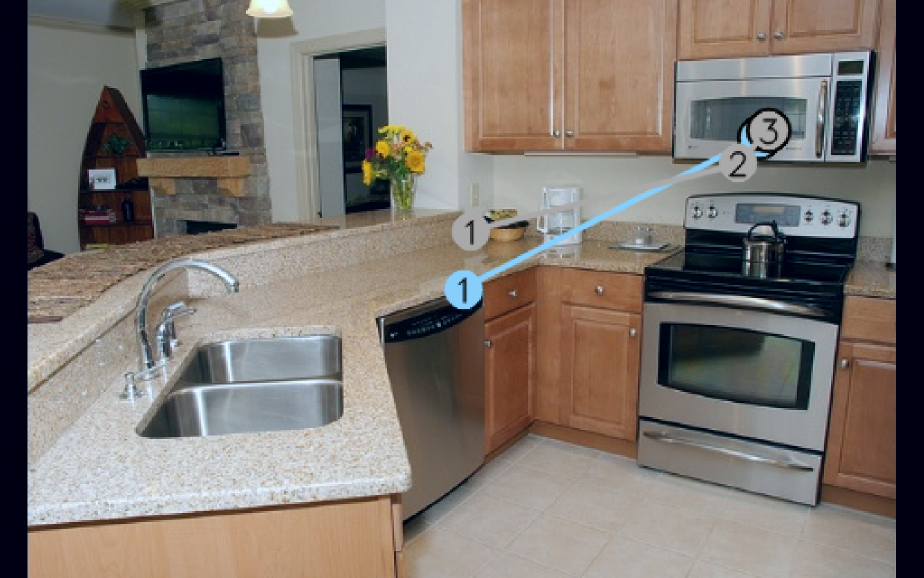}} \\ \\
        \centering \textbf{ES$_{1,5}$ } & 
        \raisebox{-0.5\height}{\includegraphics[width=0.22\textwidth]{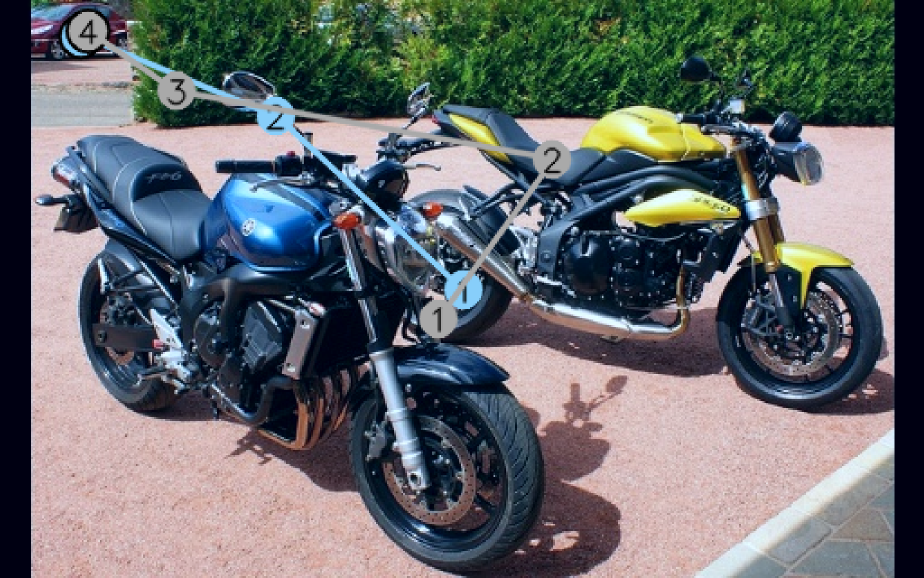}} & 
        \raisebox{-0.5\height}{\includegraphics[width=0.22\textwidth]{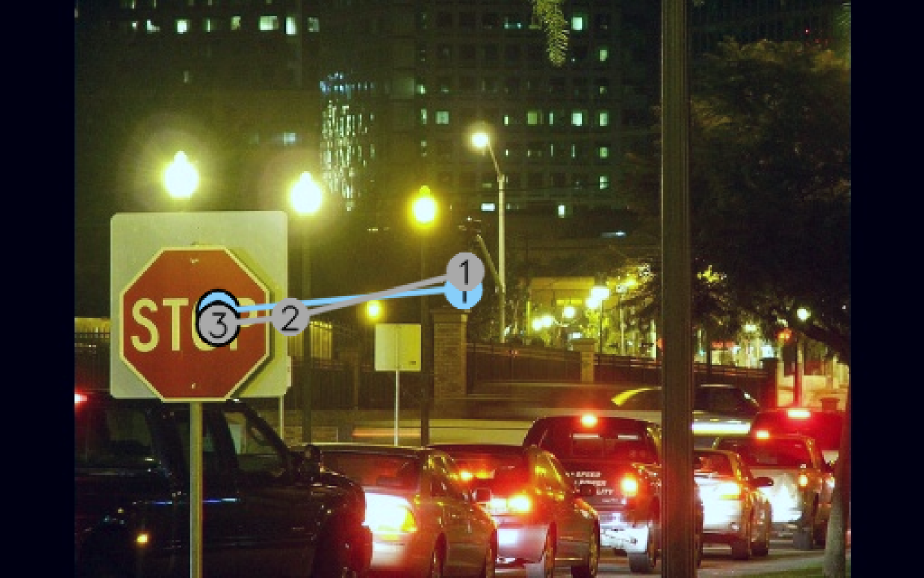}} & 
        \raisebox{-0.5\height}{\includegraphics[width=0.22\textwidth]{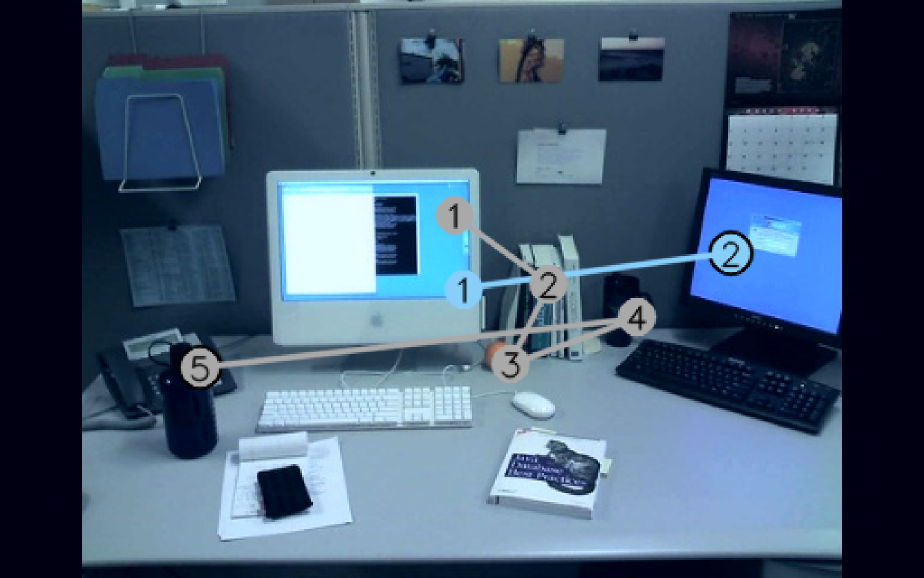}} & 
        \raisebox{-0.5\height}{\includegraphics[width=0.22\textwidth]{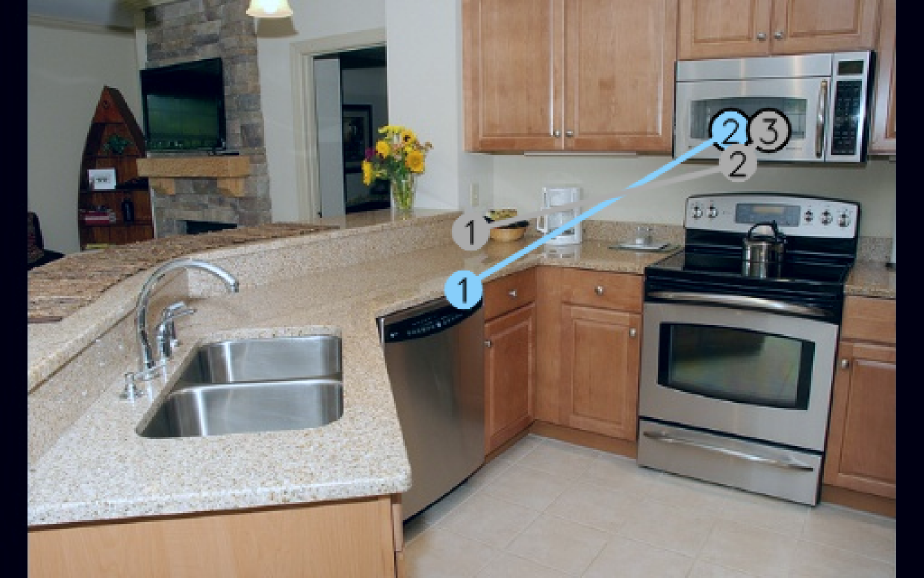}} \\ \\
        \centering \textbf{HAT} & 
        \raisebox{-0.5\height}{\includegraphics[width=0.22\textwidth]{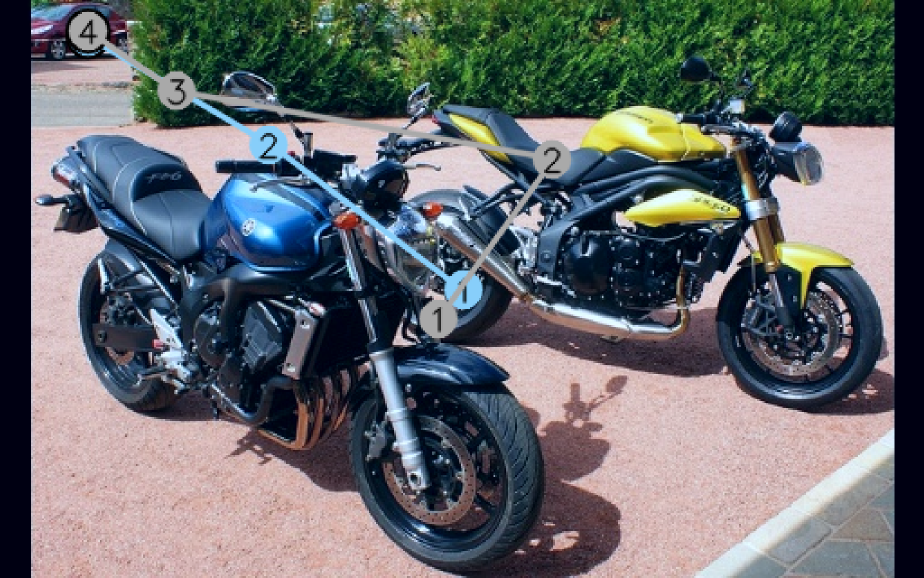}} & 
        \raisebox{-0.5\height}{\includegraphics[width=0.22\textwidth]{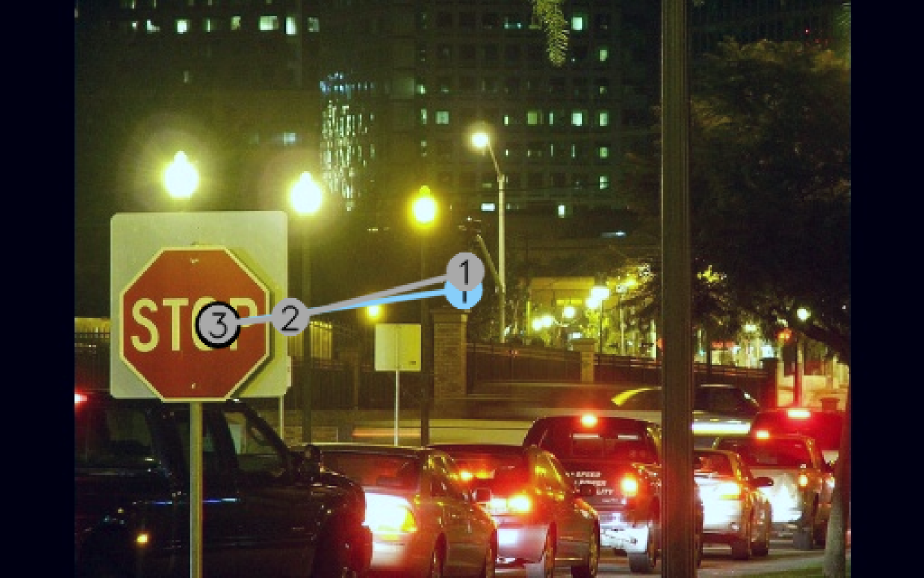}} & 
        \raisebox{-0.5\height}{\includegraphics[width=0.22\textwidth]{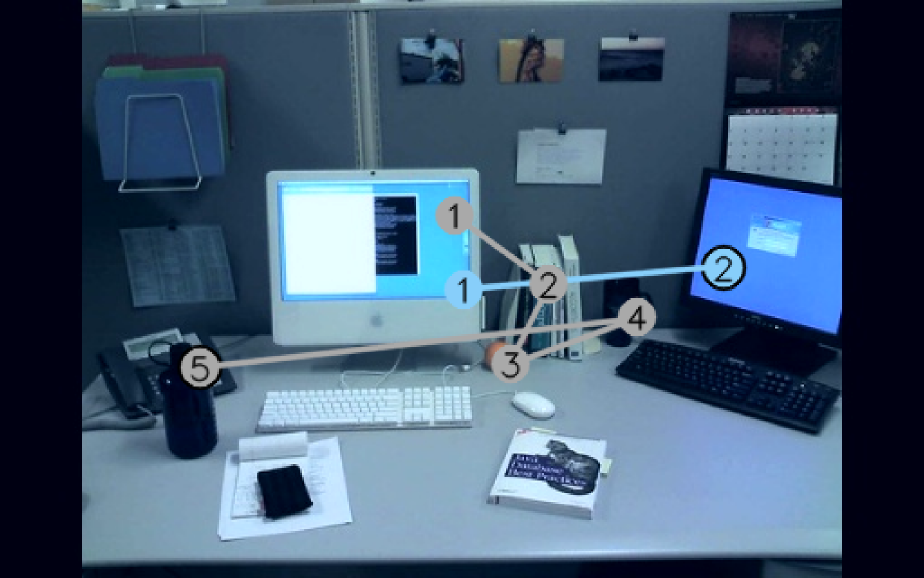}} & 
        \raisebox{-0.5\height}{\includegraphics[width=0.22\textwidth]{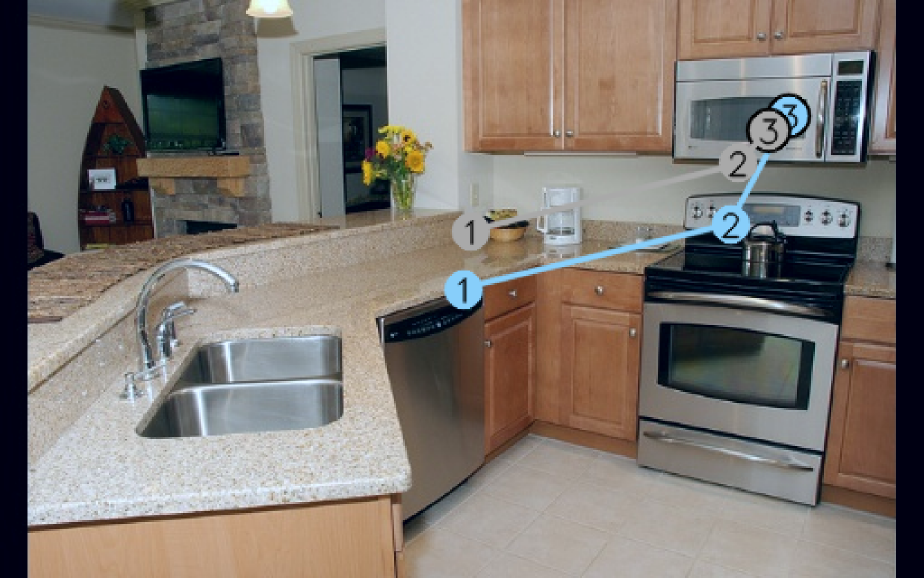}} \\ \\
    \end{tabular}
    \caption{Scanpath visualizations for the end-to-end trained HAT model on visual search (bottom row) and models that reuse selected layers from the pre-trained HAT model trained on free-viewing. Each column represents a different search target. Predicted scanpaths are shown in blue, and ground truth scanpaths are shown in brown. The final fixation in each case is highlighted with a black border.}
    \label{fig: scanpath visualization for model using pretrained HAT}
\end{figure*}

To enable qualitative understanding, Figure~\ref{fig: scanpath visualization for model using pretrained HAT} presents a visualization of the predicted scanpaths for each shared configuration, as well as the HAT model trained end-to-end on the visual search task, compared against the ground truth human scanpaths. As illustrated, despite relying on pretrained layers from the free-viewing task, the proposed models are able to produce scanpaths that closely resemble those generated by the fully trained HAT model (specific to visual search) for most target objects. Notably, in the case of the bottle target, the HAT model struggles to locate the object and generates an incorrect scanpath. In contrast, one of the models reusing the free-viewing shared layers is able to correctly locate the bottle target. This improvement may demonstrate that relying on shared layers attuned to free-viewing can at least in certain scenarios lead to accuracy benefits within the visual search task, possibly by encouraging the model to learn a broader and more diverse set of visual features. 

\begin{figure*}[!h]
    \centering
    \setlength{\tabcolsep}{4pt} 
    \begin{tabular}{ccccc}
        & \textbf{Clock} & \textbf{TV} & \textbf{Sink} & \textbf{Chair} \\ 
        \centering \textbf{LS} & 
        \raisebox{-0.5\height}{\includegraphics[width=0.22\textwidth]{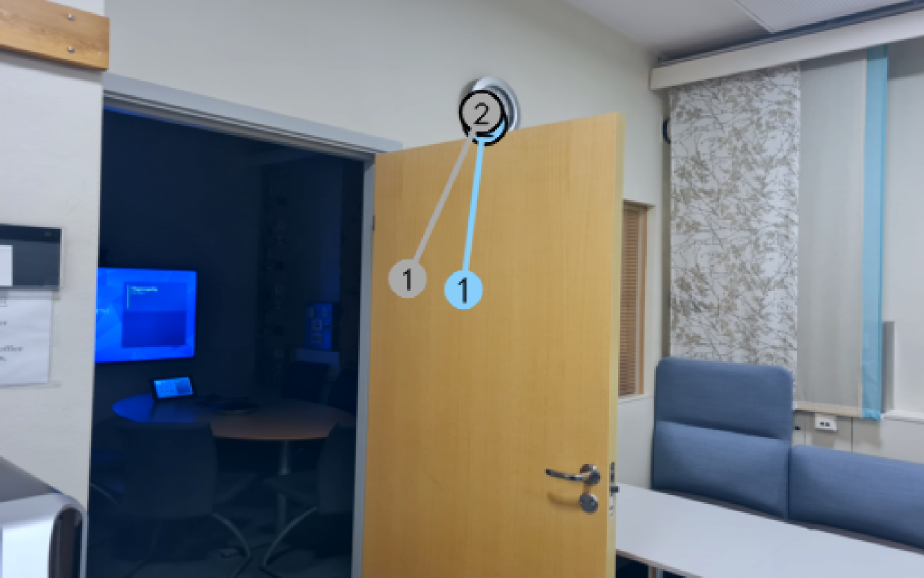}} & 
        \raisebox{-0.5\height}{\includegraphics[width=0.22\textwidth]{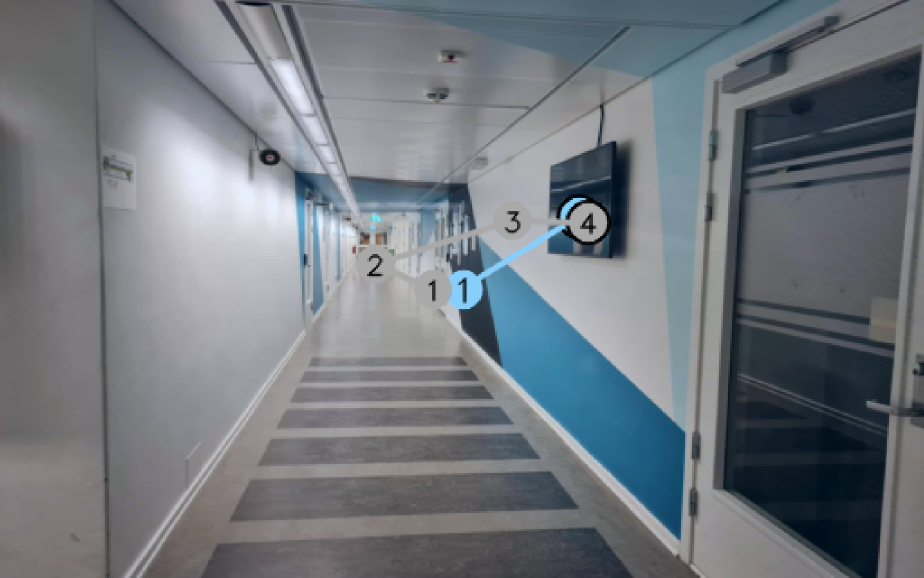}} & 
        \raisebox{-0.5\height}{\includegraphics[width=0.22\textwidth]{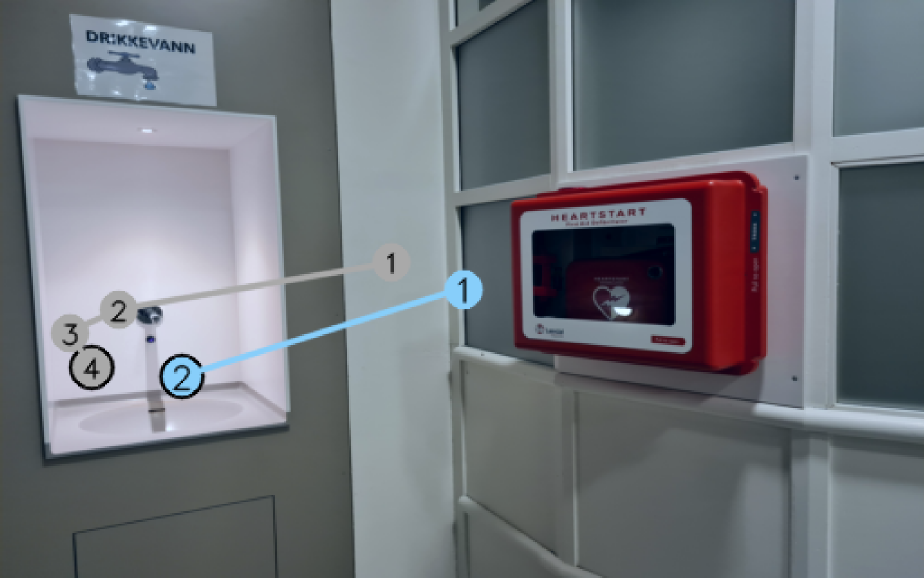}} & 
        \raisebox{-0.5\height}{\includegraphics[width=0.22\textwidth]{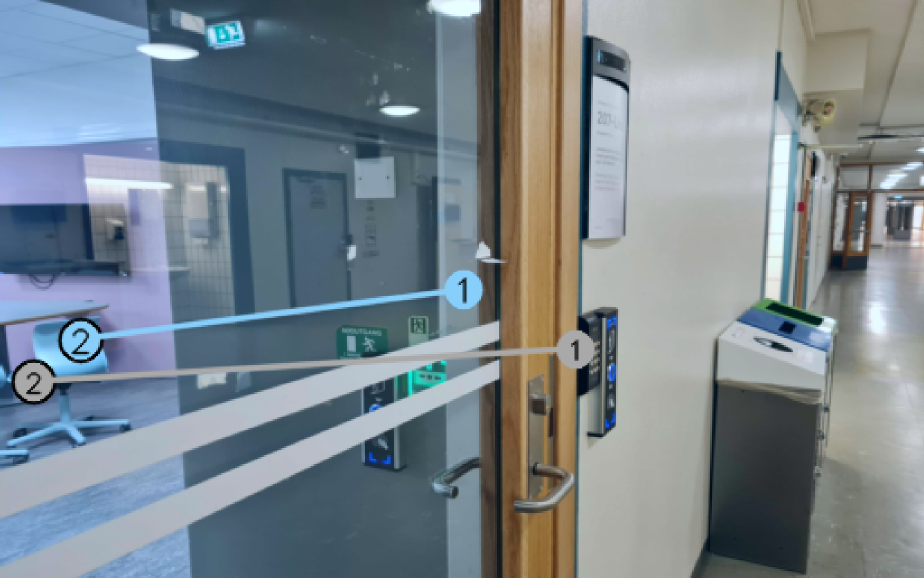}} \\ \\
        \centering \textbf{ES$_{5,1}$ }& 
        \raisebox{-0.5\height}{\includegraphics[width=0.22\textwidth]{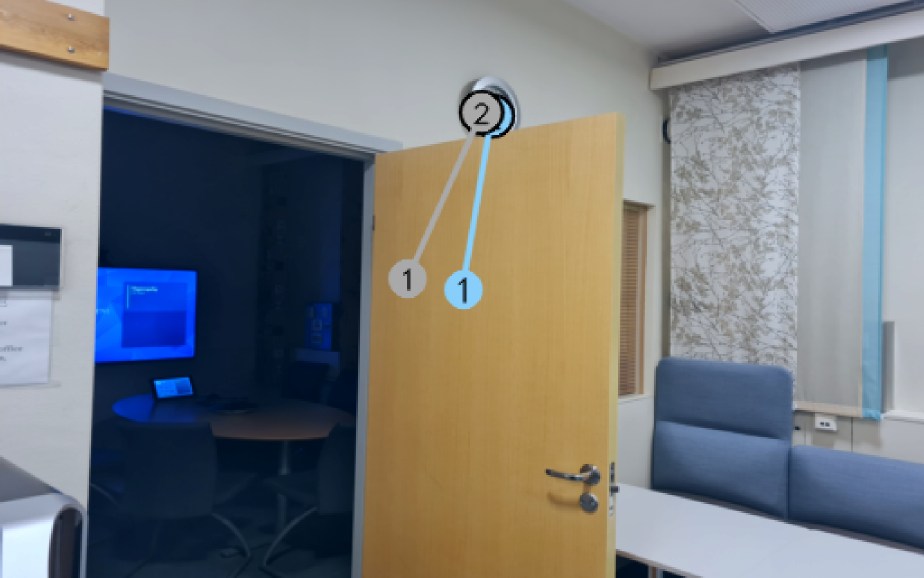}} & 
        \raisebox{-0.5\height}{\includegraphics[width=0.22\textwidth]{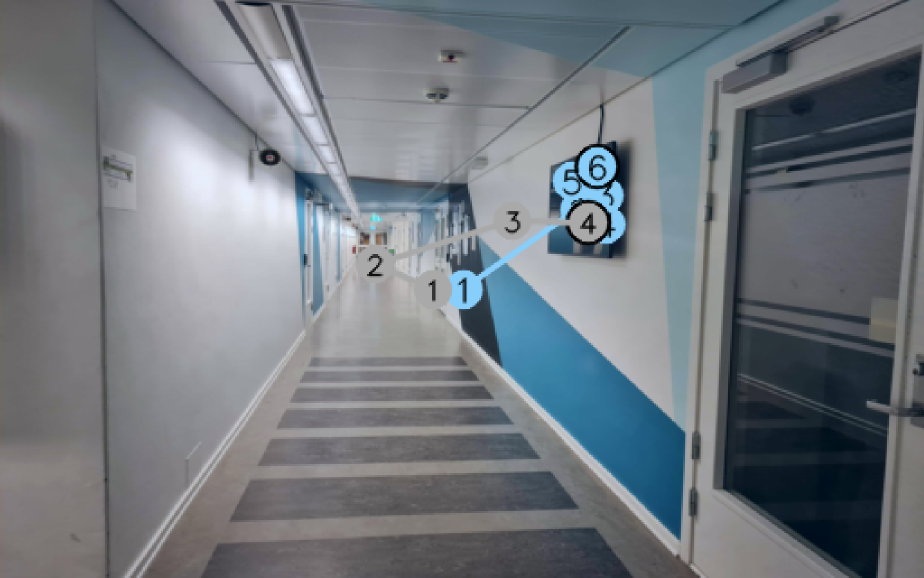}} & 
        \raisebox{-0.5\height}{\includegraphics[width=0.22\textwidth]{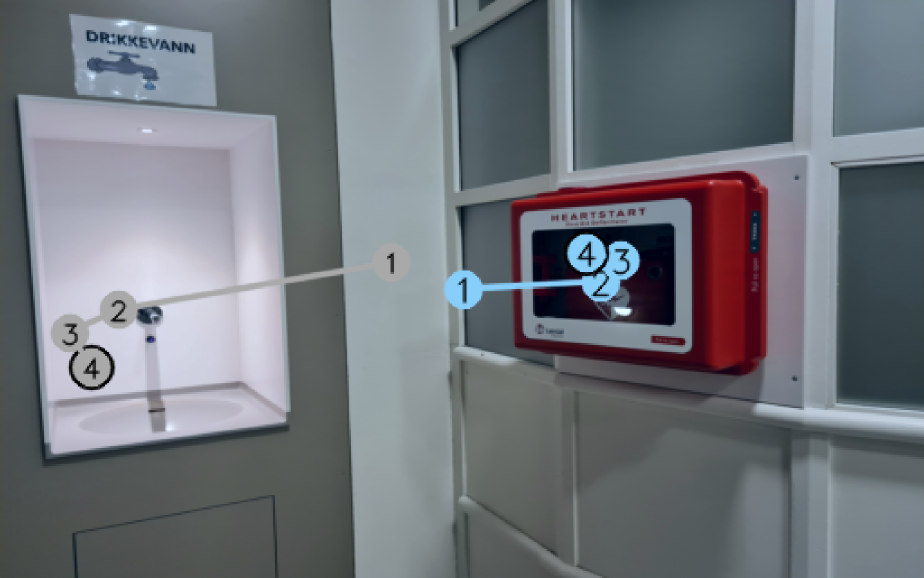}} & 
        \raisebox{-0.5\height}{\includegraphics[width=0.22\textwidth]{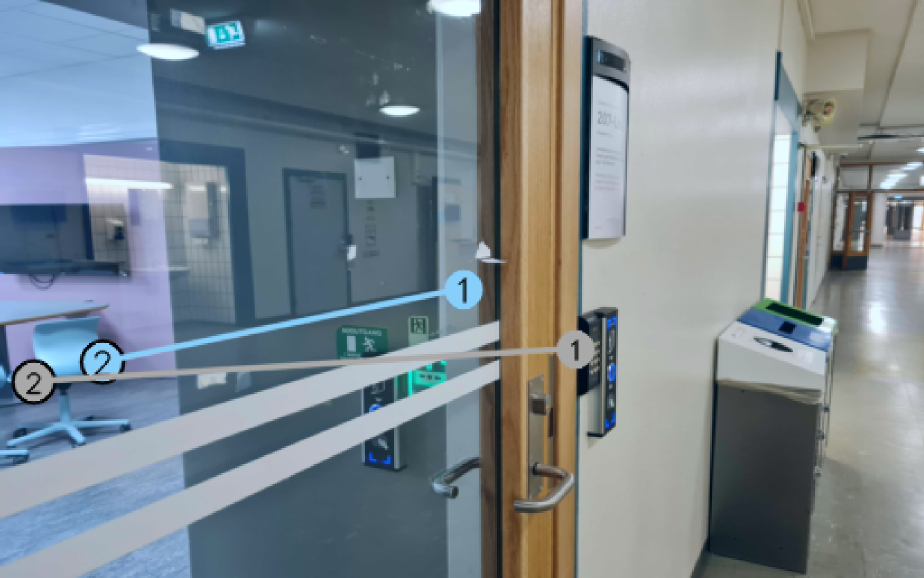}} \\ \\
        \centering \textbf{ES$_{4,2}$ } & 
        \raisebox{-0.5\height}{\includegraphics[width=0.22\textwidth]{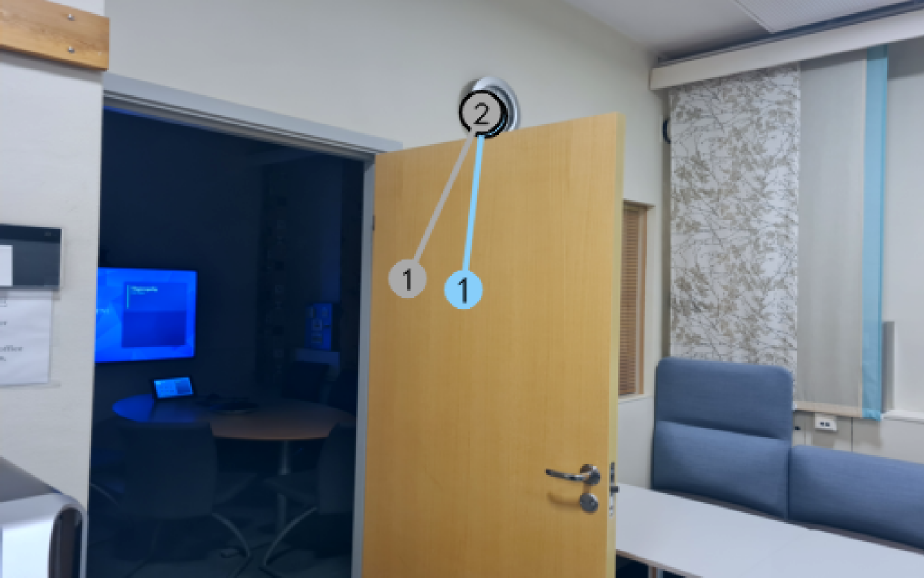}} & 
        \raisebox{-0.5\height}{\includegraphics[width=0.22\textwidth]{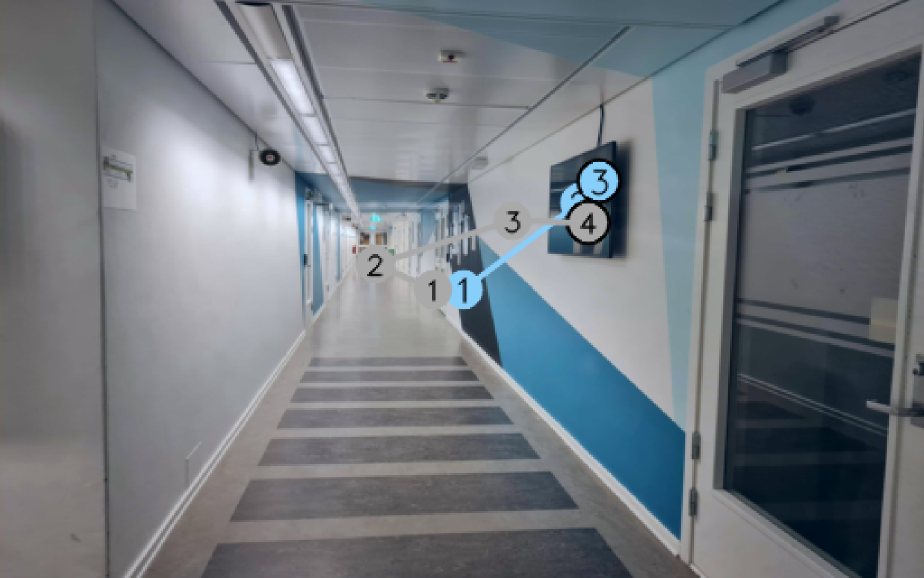}} & 
        \raisebox{-0.5\height}{\includegraphics[width=0.22\textwidth]{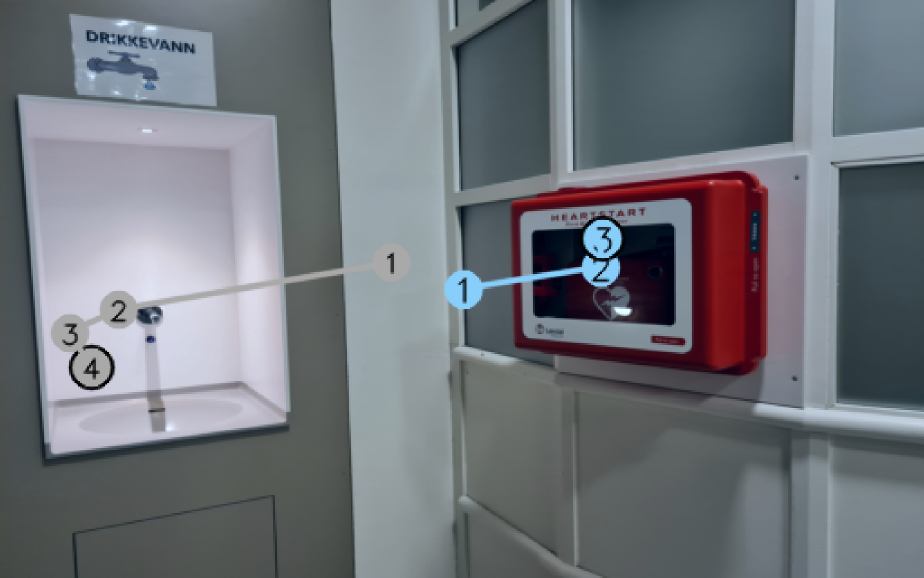}} & 
        \raisebox{-0.5\height}{\includegraphics[width=0.22\textwidth]{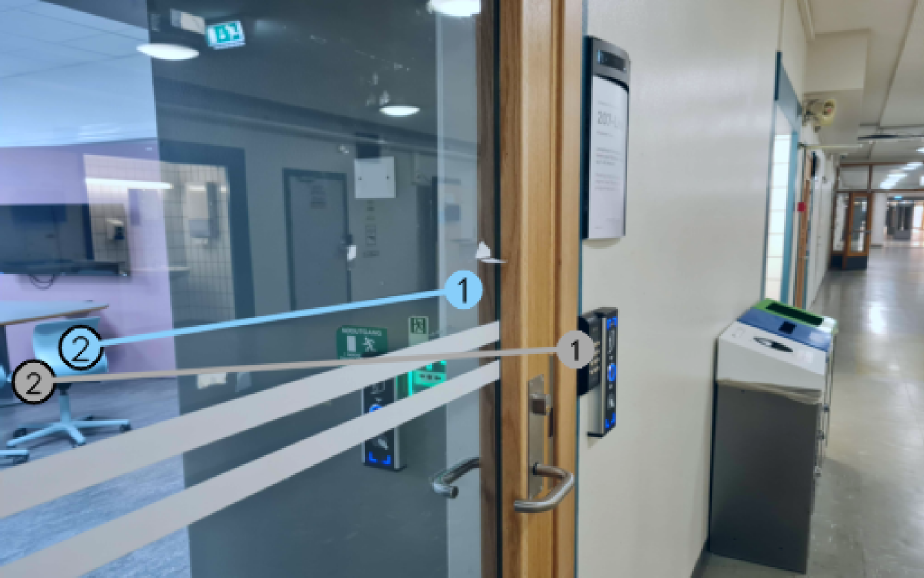}} \\ \\
        \centering \textbf{ES$_{3,3}$ } & 
        \raisebox{-0.5\height}{\includegraphics[width=0.22\textwidth]{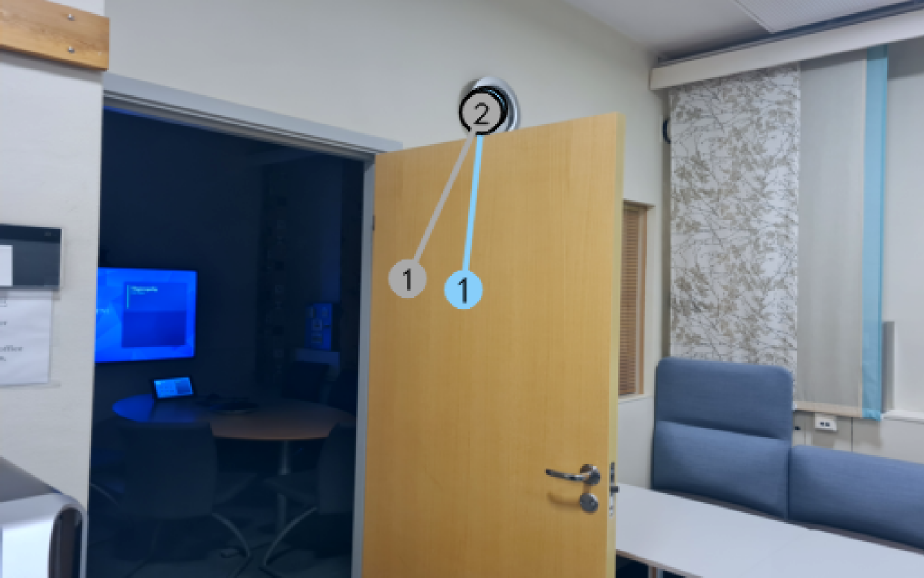}} & 
        \raisebox{-0.5\height}{\includegraphics[width=0.22\textwidth]{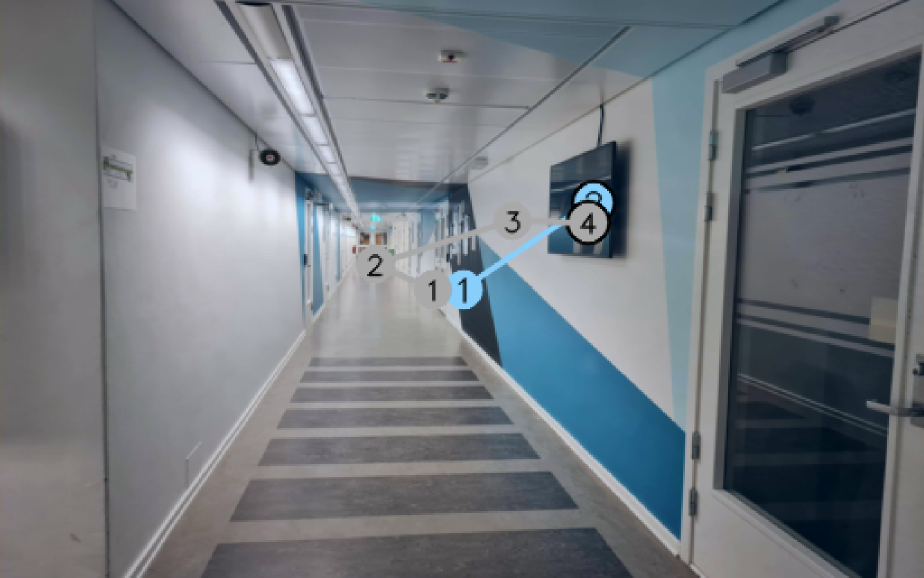}} & 
        \raisebox{-0.5\height}{\includegraphics[width=0.22\textwidth]{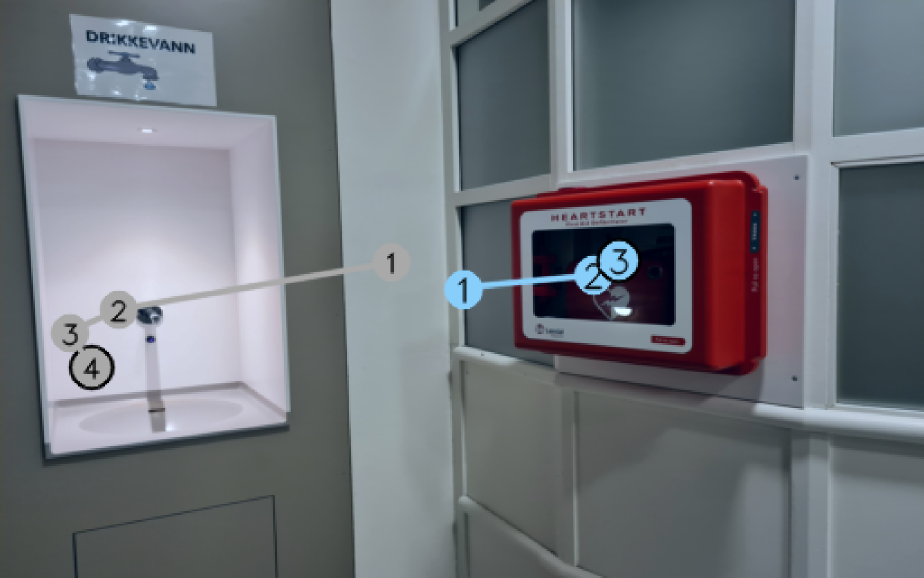}} & 
        \raisebox{-0.5\height}{\includegraphics[width=0.22\textwidth]{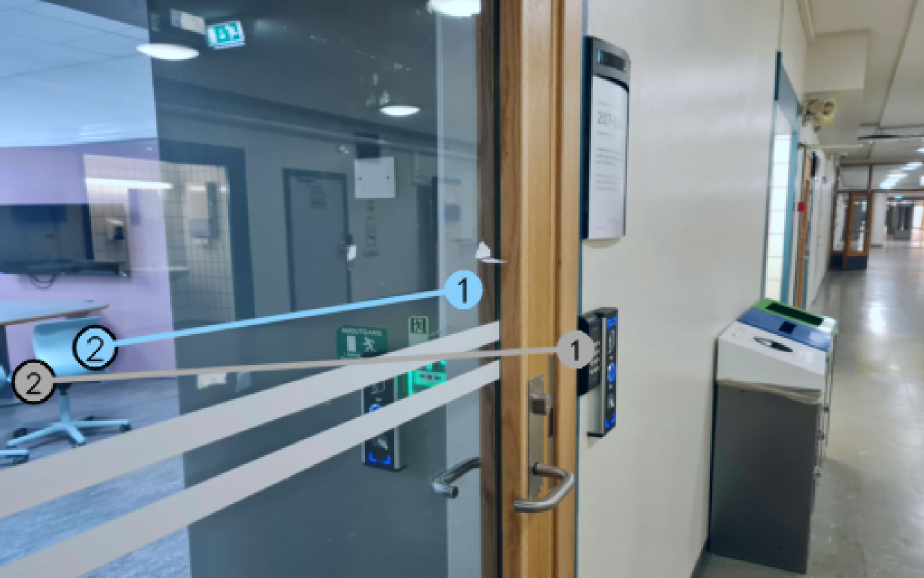}} \\ \\
        \centering \textbf{ES$_{2,4}$ } & 
        \raisebox{-0.5\height}{\includegraphics[width=0.22\textwidth]{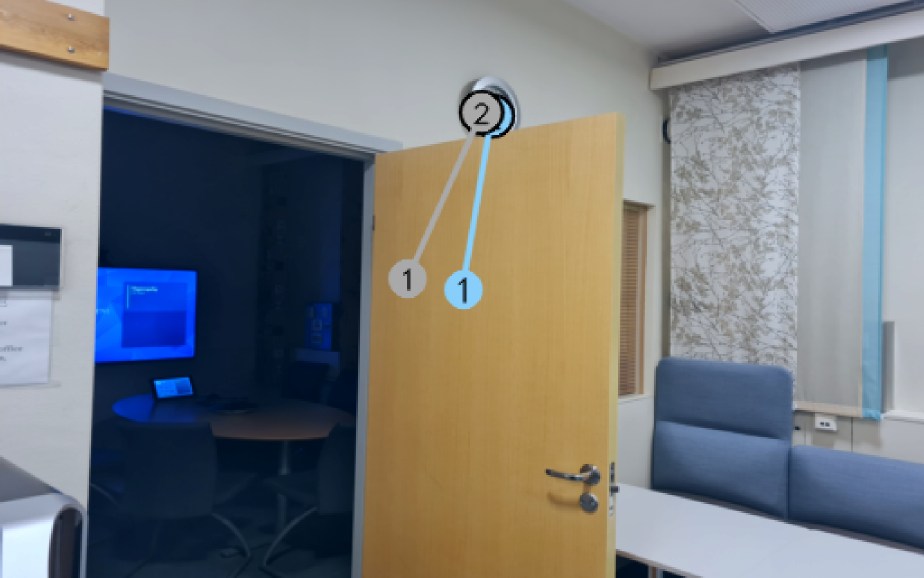}} & 
        \raisebox{-0.5\height}{\includegraphics[width=0.22\textwidth]{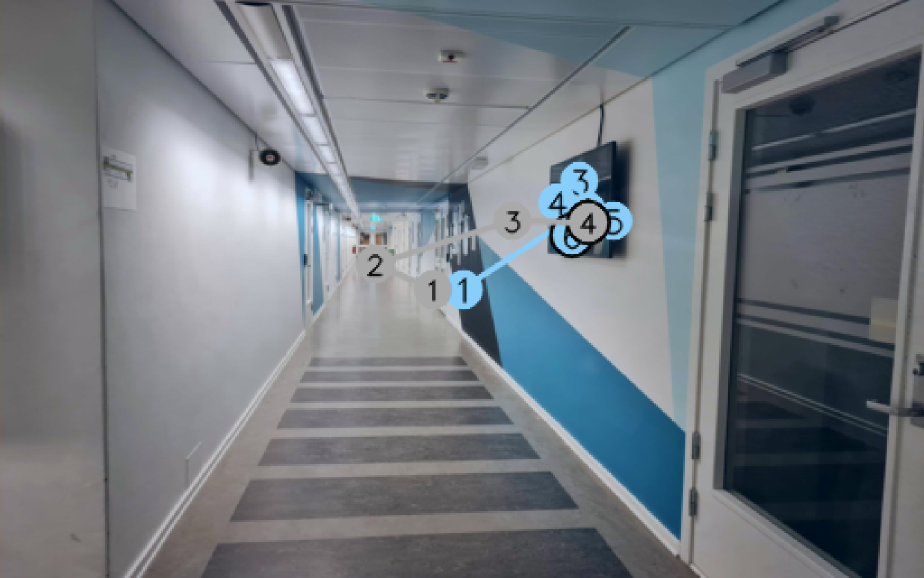}}& 
        \raisebox{-0.5\height}{\includegraphics[width=0.22\textwidth]{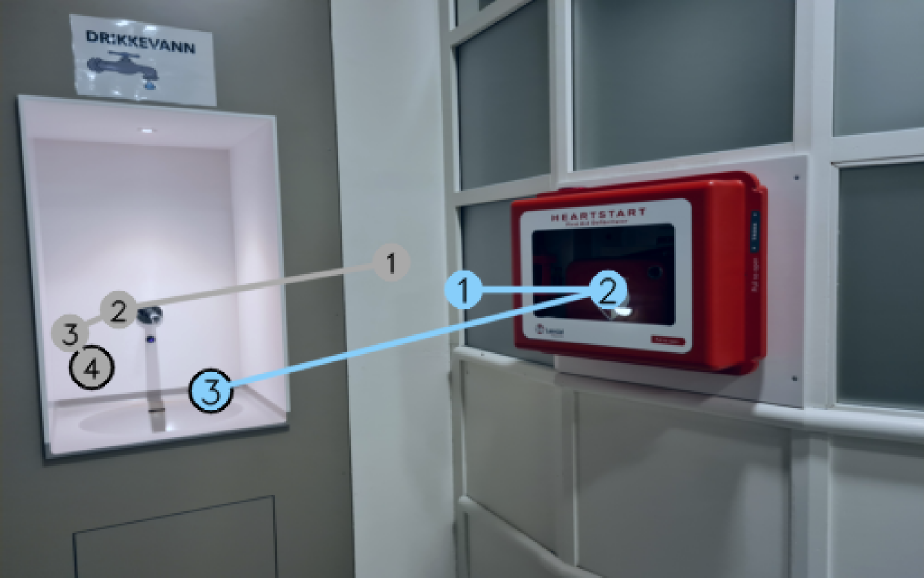}} & 
        \raisebox{-0.5\height}{\includegraphics[width=0.22\textwidth]{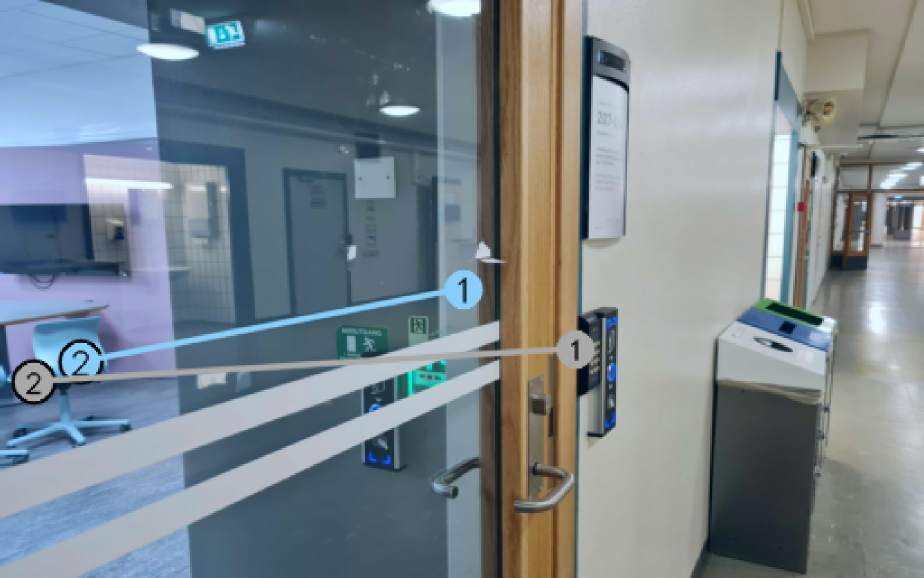}} \\ \\
        \centering \textbf{ES$_{1,5}$ } & 
        \raisebox{-0.5\height}{\includegraphics[width=0.22\textwidth]{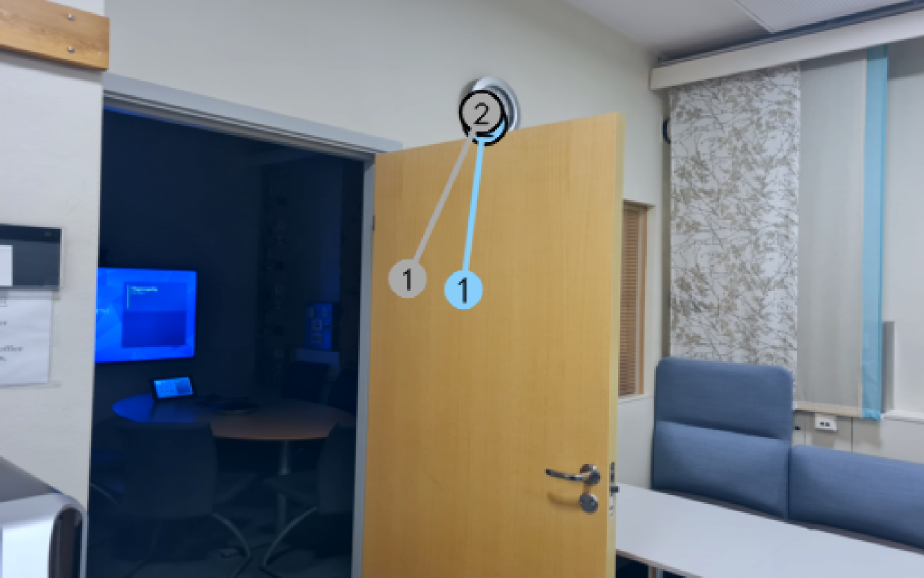}} & 
        \raisebox{-0.5\height}{\includegraphics[width=0.22\textwidth]{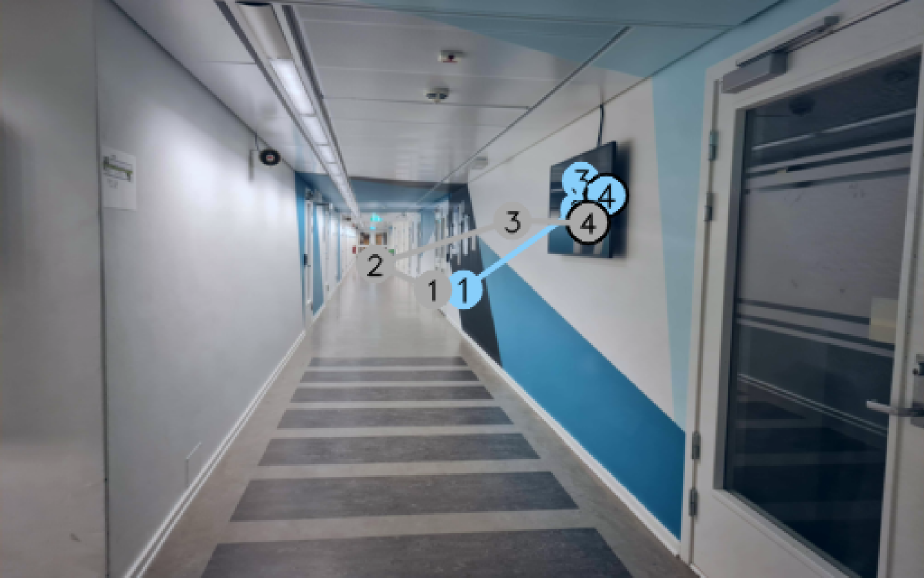}} & 
        \raisebox{-0.5\height}{\includegraphics[width=0.22\textwidth]{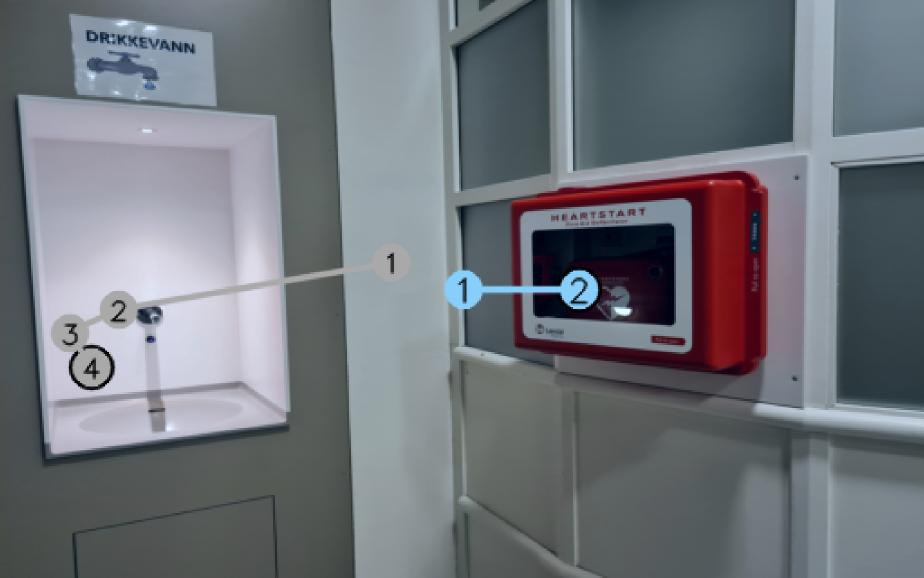}} & 
        \raisebox{-0.5\height}{\includegraphics[width=0.22\textwidth]{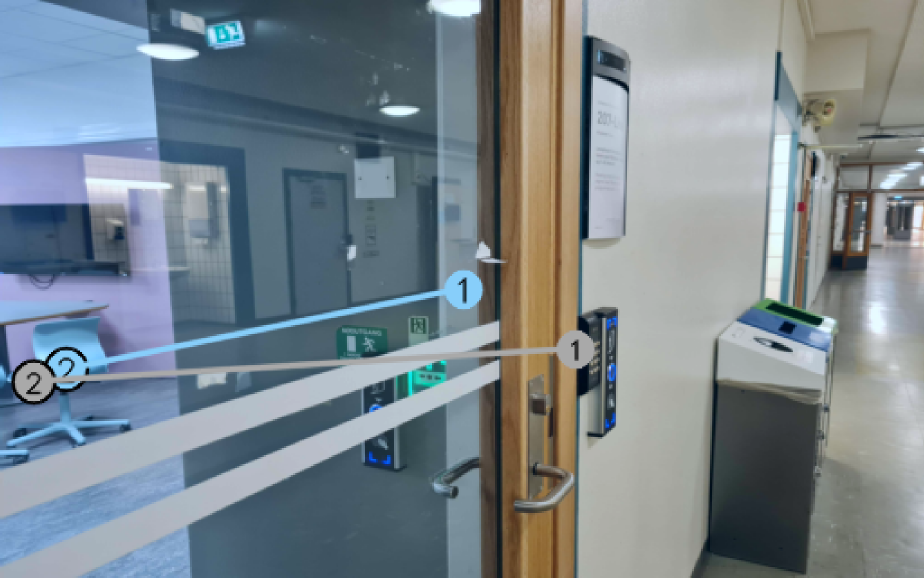}} \\ \\
        \centering \textbf{HAT} & 
        \raisebox{-0.5\height}{\includegraphics[width=0.22\textwidth]{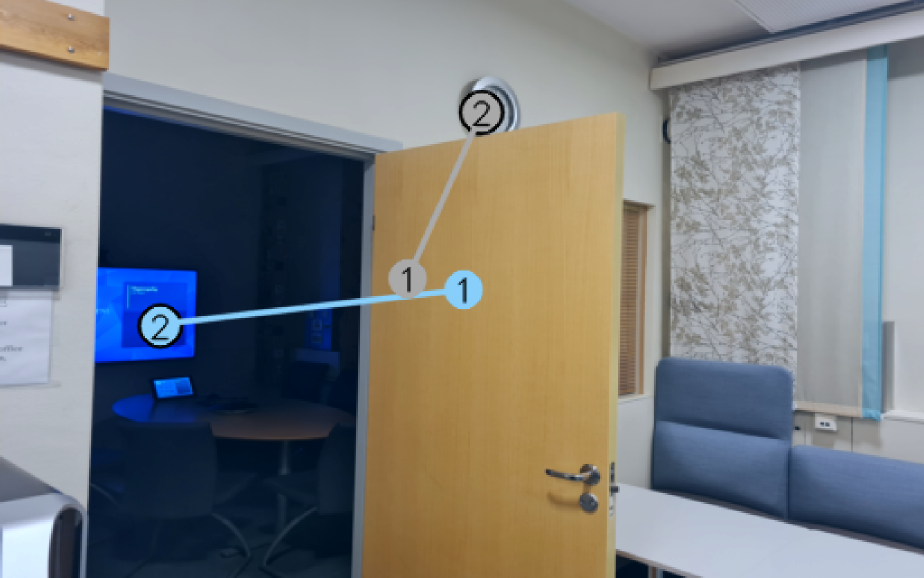}} & 
        \raisebox{-0.5\height}{\includegraphics[width=0.22\textwidth]{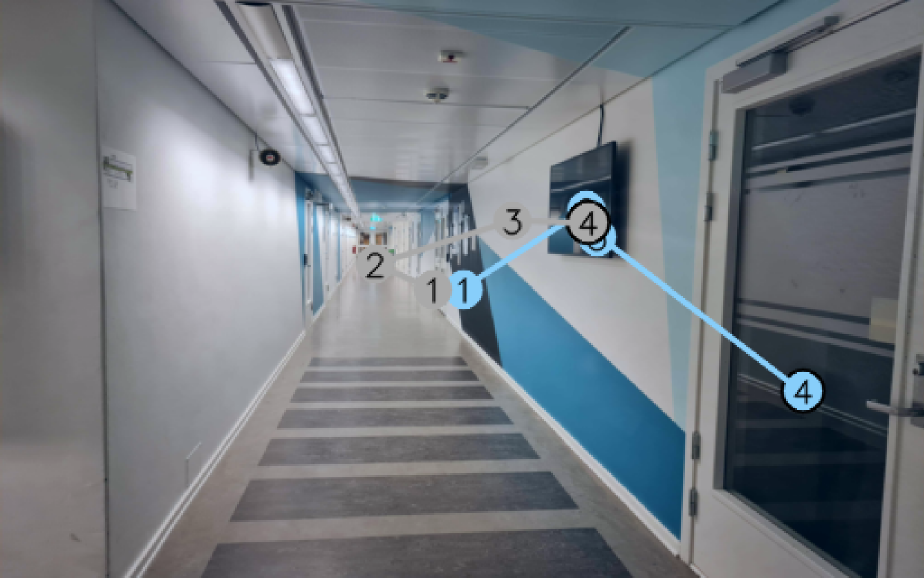}} & 
        \raisebox{-0.5\height}{\includegraphics[width=0.22\textwidth]{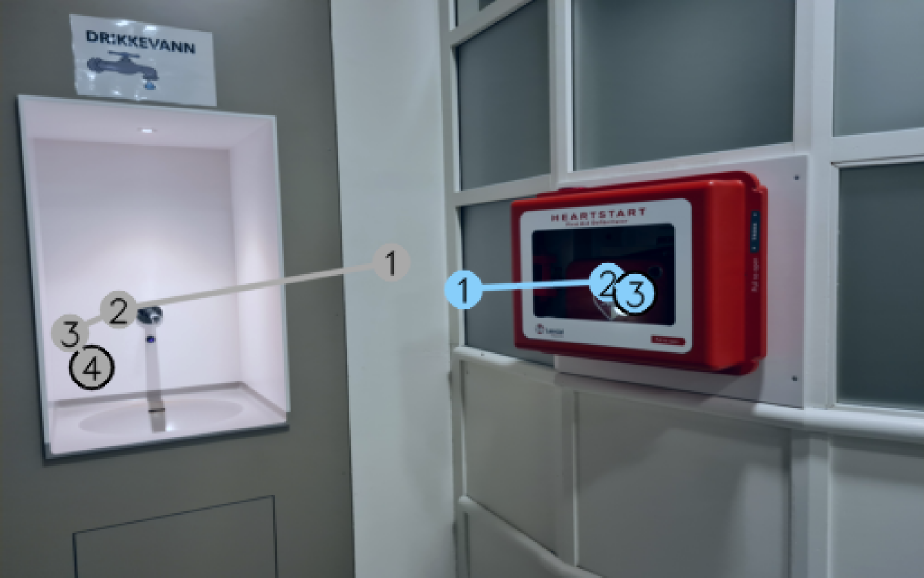}} & 
        \raisebox{-0.5\height}{\includegraphics[width=0.22\textwidth]{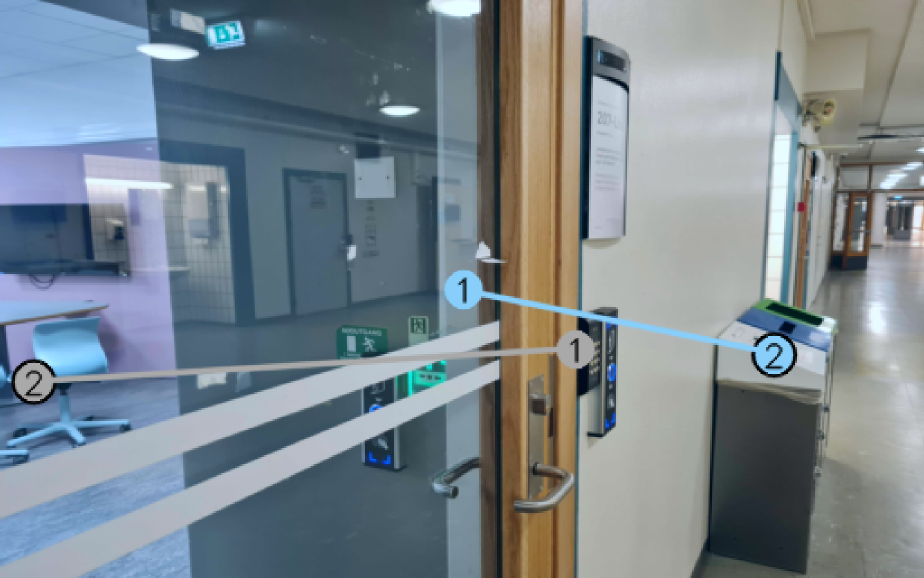}} \\ \\
    \end{tabular}
    \caption{Scanpath visualizations on the newly collected dataset for the end-to-end trained HAT model on visual search (bottom row) and models that reuse selected layers from the pre-trained HAT model trained on free-viewing. Each column represents a different search target. Predicted scanpaths are shown in blue, and ground truth scanpaths are shown in brown. The final fixation in each case is highlighted with a black border.}
    \label{fig: scanpath visualization for model using pretrained HAT (new collected data)}
\end{figure*}

\subsection{Generalization to unseen data}
To further evaluate the generalization capability of models that reuse layers from the free-viewing task, an additional dataset was collected. The newly collected dataset has 82 images, mostly in an office-like environment, where each image contains one or more search target objects. This results in 96 fixation scanpaths. The dataset contains 12 objects from the 18 objects present in the COCO-Search18 dataset but with different scenes not present in the COCO-Search18 dataset. Fixations were recorded using the Neon eye tracking glasses~\cite{baumann2023neon} by one subject.

A comparison between the performance of the proposed approach that reuses several layers from free-viewing and the HAT model on the newly collected dataset is shown in Table~\ref{tab: comparison on the new dataset between HAT and our model using pretrained HAT weights}. Here, the conditional saliency metrics are only computed due to the unavailability of the clustering method used in HAT for computing SS and the lack of segmentation annotations for SemSS on the new dataset.

The results on this unseen data show the ability of the models reusing free-viewing layers to generalize to unseen data (in most cases) more effectively than the end-to-end trained HAT model on visual search.

Figure~\ref{fig: scanpath visualization for model using pretrained HAT (new collected data)} highlights some cases of correct scanpaths predicted by variants of the proposed approach compared to incorrect ones by HAT. The results indicate that reusing some layers pretrained on free-viewing not only reduces the training computational cost significantly but also leads to better generalization to unseen data.

\section{Conclusion}\label{sec: Conclusions}
This work explored the existence of a shared representation between free-viewing and visual search tasks using the HAT~\cite{yang2024unifying} architecture. The architecture of the HAT model was modified to support both shared layers and task-specific layers, enabling an investigation into how much of the network can be reused across both tasks and the extent to which earlier processing steps tailored to free-viewing can be repurposed for task-specific attention. 

The results indicate that a common representation exists between free-viewing and visual search, enabling reusing layers from the free-viewing task for visual search attention prediction, while achieving performance comparable with that of the HAT model when trained ``end-to-end'' (except of its ResNet-50 based pixel encoder) on visual search. 

These findings lead to significant reductions in both computational cost, by 92.29\% GFLOPs and model size by 31.23\% trainable parameters. 

The ability of the HAT model to maintain, especially through the revisions and strategies proposed in this work, strong performance even when some layers are frozen from the free-viewing task may highlight its potential suitability for more complex tasks beyond visual search. In future work, we plan to incorporate an additional task into this framework to explore shared representations across more than two tasks, further expanding the versatility and efficiency of multi-task learning using shared features.
\bibliographystyle{IEEEtran} 
\bibliography{references}
\end{document}